\documentclass[pdflatex,sn-apa]{sn-jnl}% APA Reference Style
\usepackage{graphicx}%
\usepackage{multirow}%
\usepackage{amsmath,amssymb,amsfonts}%
\usepackage{amsthm}%
\usepackage{mathrsfs}%
\usepackage[title]{appendix}%
\usepackage{xcolor}%
\usepackage{textcomp}%
\usepackage{manyfoot}%
\usepackage{booktabs}%
\usepackage{algorithm}%
\usepackage{algorithmicx}%
\usepackage{algpseudocode}%
\usepackage{listings}%

\newcommand{\grayrow}[7]{#1 & #2 & #3 & #4 & #5 & #6 & #7}
\newcommand{\citepp}[1]{}
\usepackage{comment}
\usepackage{colortbl}
\usepackage[table]{xcolor}
\usepackage{booktabs}
\usepackage{makecell}
\usepackage{diagbox}
\usepackage{tabularx,array} % in the preamble

\newcolumntype{Y}{>{\centering\arraybackslash}X}

\theoremstyle{thmstyleone}%
\theoremstyle{thmstyletwo}%

\theoremstyle{thmstylethree}%

\newcommand{\ta}[1]{{#1}}
\usepackage{eqparbox}
\newdimen{\algindent}
\algnewcommand\LeftComment[2]{%
\hspace{#1\algindent}$\triangleright$ \eqparbox{COMMENT}{\textbf{\textit{#2}}} \hfill %
}

\begin{document}

\title[Article Title]{SAMix: Calibrated and Accurate Continual Learning via Sphere-Adaptive Mixup and Neural Collapse}


\author*[1]{\fnm{Trung-Anh} \sur{Dang}}\email{trung-anh.dang@univ-orleans.fr}

\author[1]{\fnm{Vincent} \sur{Nguyen}}\email{vincent.nguyen@univ-orleans.fr}

\author[2]{\fnm{Ngoc-Son} \sur{Vu}}\email{son.vu@utt.fr}

\author[1]{\fnm{Christel} \sur{Vrain}}\email{christel.vrain@univ-orleans.fr}

\affil*[1]{\orgdiv{LIFO UR 4022}, \orgname{Université d'Orléans, INSA CVL}, \orgaddress{\city{Orléans}, \postcode{45067}, \country{France}}}

\affil[2]{\orgdiv{LIST3N}, \orgname{Université de Technologie de Troyes}, \orgaddress{\city{Troyes}, \postcode{10010}, \country{France}}}

% \affil[3]{\orgdiv{Department}, \orgname{Organization}, \orgaddress{\street{Street}, \city{City}, \postcode{610101}, \state{State}, \country{Country}}}


\abstract{
While most continual learning methods focus on mitigating forgetting and improving accuracy, they often overlook the critical aspect of network calibration, despite its importance. Neural collapse, a phenomenon where last-layer features collapse to their class means, has demonstrated advantages in continual learning by reducing feature-classifier misalignment. Few works aim to improve the calibration of continual models for more reliable predictions. Our work goes a step further by proposing a novel method that not only enhances calibration but also improves performance by reducing overconfidence, mitigating forgetting, and increasing accuracy. We introduce Sphere-Adaptive Mixup (SAMix), an adaptive mixup strategy tailored for neural collapse-based methods. SAMix adapts the mixing process to the geometric properties of feature spaces under neural collapse, ensuring more robust regularization and alignment. Experiments show that SAMix significantly boosts performance, surpassing SOTA methods in continual learning while also improving model calibration. SAMix enhances both across-task accuracy and the broader reliability of predictions, making it a promising advancement for robust continual learning systems.
}




\maketitle

\section{Introduction}\label{sec1}
To achieve human-like lifelong learning, deep neural networks (DNNs) must support continual learning (CL), the ability to acquire new knowledge incrementally while retaining previously learned information. However, a major challenge is catastrophic forgetting \citep{mccloskey_1989, robins1995_catastrophic_fr}, where new learning disrupts old knowledge.
% leading to a drastic performance drop on earlier tasks. 
In response, CL has rapidly evolved with numerous approaches \citep{rusu_2016, co2l, dsdm_eccv_2022, wen2024provablecontrastivecontinuallearning, tang_kaizen_wacv_2024, anh_wacv_2025}, primarily aimed at balancing adaptation to new data (plasticity) and retention of past knowledge (stability). Compared to joint training \citep{yoon_2018, buzzega2020darkexperiencegeneralcontinual, tiwari2022gcrgradientcoresetbased}, decoupling representation and classifier learning has shown superior performance in both continual supervised \citep{co2l, wen2024provablecontrastivecontinuallearning, li_cl_important_sampling_2024, anh_wacv_2025} and self-supervised learning \citep{fini2022cassle, pseudo_negative_icml_2024, madaan_2022_lump, cromomix_mushtaq_eccv_2024}. Among these, \textbf{mixup} \citep{mixup_2018} has emerged as a simple yet effective technique, primarily applied in continual self-supervised learning \citep{madaan_2022_lump, cromomix_mushtaq_eccv_2024} to promote smoother task transitions and increase sample diversity, especially for old samples in buffers. Beyond reducing forgetting, mixup also acts as a \textbf{calibration} method, aligning predicted confidence scores with actual accuracy\ta{, as demonstrated in standard supervised learning and further extended to static imbalanced and long-tailed settings by ReMix~\citep{remix_2021}}. In DNNs, calibration reflects prediction reliability - a crucial factor for safety-sensitive applications that often require CL, such as medical diagnosis \citep{medical_xiaoqian_2012} and autonomous driving \citep{autonomous_driving_2020}, where unreliable predictions can pose serious risks. \ta{Moreover, recent analyses in class-imbalanced settings indicate that the reliability of predicted probabilities is highly sensitive to distributional assumptions and resampling-based class rebalancing mechanisms \citep{PICCININNI2024104666, carvalhoResamplingApproachesHandle2025}. Since CL naturally induces class imbalance between current and past tasks, this sensitivity poses additional challenges for maintaining reliable calibration under sequential data and evolving representations.} However, the role of mixup in mitigating overconfidence and enhancing CL performance remains underexplored.

% However, the role of mixup in mitigating overconfidence and enhancing CL performance remains underexplored.

\textbf{Neural collapse (NC)}, where last-layer features converge to their class means during training, has attracted significant attention in DNNs across both \textit{static} and \textit{dynamic} (continual) learning scenarios \citep{lu2021neuralcollapsecrossentropyloss, fang_pnas_2021, wenlong_ji_iclr_2022, zhou_icml_2022, fisher2024pushingboundariesmixupsinfluence}.
Most NC-based methods rely on the dot-regression loss (DR) \citep{yang_induce_nc_2022}, which enforces feature alignment with fixed class prototypes to enhance class separation and reduce misalignment between features and classifiers. Recently, focal neural collapse contrastive (FNC$^2$) \citep{anh_wacv_2025} loss was introduced to leverage NC more flexibly in CL. Unlike DR, which rigidly aligns features with class prototypes, FNC$^2$ optimizes feature relationships, enhancing intra-class diversity and improving generalization. While NC-based methods effectively improve accuracy and mitigate forgetting, their impact on calibration remains unexplored, as does the effect of calibration in NC-based methods on overall model performance. \ta{Indeed, recent studies have highlighted the importance of calibration in CL beyond accuracy alone \citep{li2024calibrationcontinuallearningmodels, hwang2025tciltemperaturescalingusing}. In particular, \citep{li2024calibrationcontinuallearningmodels} shows that CL models are not naturally calibrated under non-stationary training compared to joint training, highlighting that forgetting mitigation alone does not ensure calibration. However, existing calibration approaches in CL remain limited in scope, primarily relying on post-hoc, global logit-scaling strategies and typically assuming uniform calibration across tasks. As a result, they do not explicitly account for key CL-specific challenges such as task-wise distribution shift, class imbalance between old and new tasks, or representation drift as learning progresses, leaving substantial room for improvement.}

% Indeed, as shown in \citep{li2024calibrationcontinuallearningmodels, hwang2025tciltemperaturescalingusing}, beyond performance, CL models must also produce well-calibrated predictions. However, existing attempts at calibration in CL remain relatively simple and leave much room for improvement.

In this work, we address both across-task accuracy and the overlooked challenge of calibration in continual learners. We conduct a comparative analysis of NC-based methods, examining their ability to balance performance and reliability in CL. Our main contributions are summarized as:

\begin{enumerate}
    \item 
    We propose Sphere-Adaptive Mixup (SAMix), a simple yet effective method to improve calibration, mitigate forgetting, and boost accuracy in NC-based CL. SAMix leverages spherical linear interpolation (Slerp) of fixed ETF (Equiangular Tight Frame) prototypes to generate adaptive mixed prototypes, enhancing representation learning and smoothing decision boundaries. 
    \footnote{We note that \citep{li2024calibrationcontinuallearningmodels, hwang2025tciltemperaturescalingusing} aims to calibrate CL models for more reliable predictions, whereas our work goes further by proposing a method that enhances calibration and also mitigates forgetting, and improves accuracy.} We also introduce new metrics to evaluate calibration and overconfidence in CL, using them to assess the reliability of NC-based methods.
    
    % \item We provide theoretical analyses to compare different NC-based methods for handling mixed samples and validate our findings through empirical results.
    \item \ta{We provide theoretical analyses to explain why Slerp is more suitable than linear interpolation for generating mixed prototypes (see Sec.~\ref{sec:phantom_method}) and to compare different NC-based plasticity losses for handling mixed samples (see Sec.~\ref{sec:considerd-nc-methods}, Appendix~\ref{sec:theoretical-model}), with empirical validation of our findings.}
    
    \item Through extensive experiments, we show that our calibration strategy significantly enhances network calibration, reduces overconfidence, and improves performance in CL. Notably, our approach with a novel method achieves SOTA results across multiple datasets, both with and without memory, in most experimental settings.
\end{enumerate}

% --------------- SEC 2 - RELATED WORK --------------
% \input{AnonymousSubmission/LaTeX/main_sec/2_relatedwork}
\section{Related Work}
\label{sec:relatedwork}

\subsection{Continual Learning}
CL algorithms can be categorized into two main paradigms: pretrained-based and training-from-scratch. Pretrained methods \citep{wang2022learningpromptcontinuallearning, wang2022dualpromptcomplementarypromptingrehearsalfree, smith2023codapromptcontinualdecomposedattentionbased, Liu_2025_CVPR} leverage large datasets for general representations, while training-from-scratch approaches suit resource-limited settings. These latter algorithms can be grouped into: \textbf{Architecture}-based methods \citep{rusu_2016, yoon_2018, li_learn_tg_2019}, which allocate task-specific parameters or expand networks; \textbf{Regularization}-based methods \citep{friedemann_si_2017, sangwon_neurips_2020, fini2022cassle, dongmin, tang_kaizen_wacv_2024, anh_wacv_2025}, which constrain weight updates to preserve past knowledge; and \textbf{Rehearsal}-based or \textbf{replay}-based methods \citep{madaan_2022_lump, buzzega2020darkexperiencegeneralcontinual, co2l, wen2024provablecontrastivecontinuallearning, cromomix_mushtaq_eccv_2024}, which replay stored data or use generative models.

A primary challenge in CL is the plasticity-stability dilemma - balancing the integration of new knowledge (plasticity) with the retention of past information (stability). Regularization-based methods often address this via knowledge distillation \citep{hinton_kd_2015}, which preserves prior knowledge during training. This may involve intermediate features \citep{hou_learning_2019} or final outputs \citep{fini2022cassle, pseudo_negative_icml_2024}. Recent advances extend distillation to relational knowledge, such as instance-wise relation distillation (IRD) in Co$^2$L \citep{co2l} and prototype-instance relation distillation (PRD) in CCLIS \citep{li_cl_important_sampling_2024}. Building on these, \citep{anh_wacv_2025} proposes hardness-softness distillation (HSD), combining IRD and a variant of PRD for fixed prototypes, achieving superior performance. In this work, we adopt HSD to preserve past knowledge through relational consistency with fixed prototypes.

\begin{figure*}[t]
    \centering
    \includegraphics[width=1.0\linewidth]{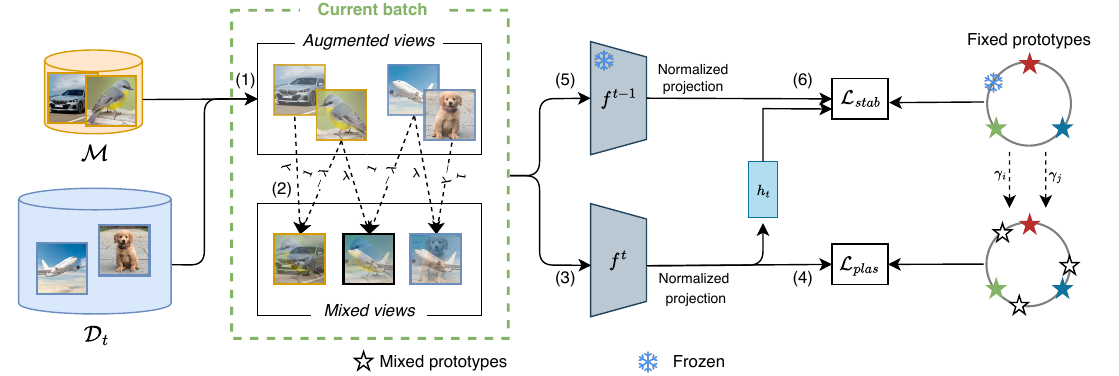}
    
    \caption{\centering Illustration of our proposed method.}
    \label{fig:overall-methods}
    % \vspace{-0.5em}
\end{figure*}

\textbf{Remarks.} While most CL methods aim to reduce forgetting and boost accuracy, they often overlook the crucial aspect of calibration. We address this gap by proposing a simple yet effective calibration method and evaluating its impact on both calibration and performance across tasks.

\subsection{Mixup}
\label{sec:mixup_related_work}
Mixup \citep{mixup_2018} is a data augmentation strategy based on the principle of vicinal risk minimization \citep{vrm_nips_2000}, which interpolates linearly between example pairs and their corresponding labels. The equation of mixup is defined as:
% {\footnotesize
\begin{equation}
    \tilde{\mathbf{x}}_{ij}=\lambda \mathbf{x}_i + (1 - \lambda) \mathbf{x}_j
    \label{eq:x_mix}
\end{equation}
% }
% {\footnotesize
\begin{equation}
     \tilde{\mathbf{y}}_{ij}=\lambda \mathbf{y}_i + (1 - \lambda) \mathbf{y}_j
    \label{eq:y_mix}
\end{equation}
% }
where $(\mathbf{x}_i, \mathbf{y}_i)$, $(\mathbf{x}_j, \mathbf{y}_j)$ are two random pairs of samples and one-hot labels drawn from the training data, and $\lambda \in [0,1]\sim  \text{Beta}(\alpha, \alpha)$, $\alpha \in (0, \infty)$.
With simplicity and ease of deployment, mixup has emerged as a well-known method, used widely in both supervised learning \citep{guo2018mixuplocallylinearoutofmanifold, verma2019manifoldmixupbetterrepresentations} and unsupervised learning \citep{kim2020mixcomixupcontrastivelearning, ren2022simpledatamixingprior, shen2022unmixrethinkingimagemixtures} to prevent overfitting and to improve generalization. Moreover, \citep{thulasidasan_on_mixup_training_2020, noh2023rankmixuprankingbasedmixuptraining} show that mixup-trained DNNs are significantly better calibrated, with softmax scores closely reflecting actual correctness likelihood.

\textbf{Mixup in CL.} Mixup has recently been used in CL by combining current task samples with memory instances to promote linearity across tasks. Most studies focus on self-supervised settings \citep{madaan_2022_lump, cromomix_mushtaq_eccv_2024}, where mixup serves both as an augmentation technique to increase the number of training samples, especially old ones in memory, and as a regularizer to distill knowledge and reduce forgetting. \ta{However, recent work \cite{Maro_as_2021} shows that mixup does not universally guarantee improved calibration and may even degrade it when linear interpolation fails to reflect the underlying data geometry or class proportion. In CL, such conditions naturally arise due to task-wise distribution shifts, imbalance between old and newly introduced classes, and representation drift across sequential updates. Therefore, naively applying linear mixup in CL may worsen confidence-accuracy alignment. These observations motivate a geometry-aware mixing strategy tailored for CL.}

\textbf{Remarks.} Like traditional mixup and its CL adaptations, we generates new samples via linear interpolation. However, instead of mixing labels, we interpolate class prototypes on a hypersphere guided by NC geometry. While mixup benefits in CL, its impact on calibration remains underexplored. This work bridges this gap by showing how our method improves calibration, reduces overconfidence, and still boosts accuracy.

\ta{Notably, SAMix aims to learn invariant representations across tasks by adopting a two-stage learning framework, where representation learning is decoupled from classifier training. The classifier is used only for evaluation, rather than being jointly optimized with the feature extractor as in one-stage learning. Therefore, existing soft-label mixup methods (e.g., standard mixup \citep{mixup_2018}, ReMix \citep{remix_2021}) are not directly applicable, as they mainly affect classifier optimization rather than representation learning. In addition, post-hoc calibration techniques (e.g., label smoothing \citep{szegedy2015rethinkinginceptionarchitecturecomputer} or temperature scaling \citep{guo2017_calibration_nn_2017}) focus only on adjusting prediction confidence and therefore are not compared in this work. Other mix-based augmentations (e.g., CutMix \citep{yun2019cutmixregularizationstrategytrain}, Manifold Mixup \citep{verma2019manifoldmixupbetterrepresentations}) operate at the input or feature level and rely on joint classifier optimization. Since SAMix performs geometry-aware mixing at the prototype level to shape representation structure, these methods are not directly comparable to our setting.}

\subsection{Neural Collapse}
\label{sec:nc_related_work}
Neural collapse (NC) \citep{papyan_2020} is a phenomenon observed during the final stage of DNN training, 
where last-layer features and classifier form a structured geometric pattern on a balanced dataset, aligning with a simplex Equiangular Tight Frame (ETF).
Building on this foundation, subsequent studies have proven NC is globally optimal in balanced training with cross-entropy \citep{lu2021neuralcollapsecrossentropyloss, zhu_neurips_2021, wenlong_ji_iclr_2022} and mean squared error \citep{poggio_explicit_ra_2020, han_nc_2021, zhou_icml_2022} losses. In addition, \citep{yang_induce_nc_2022} try to induce NC by fixing the classifier under imbalanced training, addressing the issue of misalignment between features and classifier.

The concept of NC is defined as follows.

\textbf{Definition 1.} A simplex Equiangular Tight Frame (ETF) consists of a set of $K$ vectors: $\mathbf{Q} = \{\mathbf{q}_k\}_{k=1}^{K}$, each vector $\mathbf{q}_k \in \mathbb{R}^{d}$, $K \le d+1$, which satisfies:

\begin{equation}
    \mathbf{Q} = \sqrt{\frac{K}{K-1}}\mathbf{U}\left(\mathbf{I}_K-\frac{1}{K}\mathbf{1}_K\mathbf{1}_K^T\right),
    \label{eq:etf}
\end{equation}
where $\mathbf{U} \in \mathbb{R}^{d \times K}$ is an orthogonal basis and $\mathbf{U}^{T}\mathbf{U}=\mathbf{I_K}$, $\mathbf{I}_K$ is an identity matrix and $\mathbf{1}_K$ is an all-ones vector.

Every vector $\mathbf{q}_k$ has the same $\ell_2$ norm, and the inner product between any pair of distinct vectors is given by $-\frac{1}{K-1}$, which represents the lowest possible cosine similarity for K equiangular vectors in $\mathbb{R}^d$. This configuration can be expressed as follows:
\begin{equation}
    \mathbf{q}^T_{i}\mathbf{q}_{j}=\frac{K}{K-1}\delta_{i,j}-\frac{1}{K-1},\ \ \forall i, j\in[1,K],
    \label{eq:cosine-similarity-etf}
\end{equation}
where $\delta_{i, j}=1$ in case of $i=j$, and 0 otherwise.

Subsequently, the NC phenomenon can be formally characterized by the following four attributes \citep{papyan_2020}:

NC1: The features from the final layer corresponding to the same class converge to their intra-class mean, ensuring that the covariance approaches zero, i.e., $\Sigma^{k}_V\rightarrow\mathbf{0}$. Mathematically, this is expressed as:
% Features from the last layer within the same class converge to their intra-class mean, such that the covariance $\Sigma^{k}_V\rightarrow\mathbf{0}$. Here, 
$\Sigma^{k}_V=\mathrm{Avg}_{i}\{(\boldsymbol{\nu}_{k,i}-\boldsymbol{\mu}_{k})(\boldsymbol{\nu}_{k,i}-\boldsymbol{\mu}_{k})^T\}$, where $\boldsymbol{\nu}_{k,i}$ is the feature of sample $i$ in class $k$, and $\boldsymbol{\mu}_{k}$ denotes the intra-class feature mean.

NC2: When centered by the global mean, intra-class means align with the vertices of a simplex ETF. That is, the set
% After centering by the global mean, intra-class means align with simplex ETF vertices, i.e., 
$\{\tilde{\boldsymbol{\mu}}_k\}$, $1 \le k \le K$ satisfy Eq. (\ref{eq:cosine-similarity-etf}), where $\tilde{\boldsymbol{\mu}}_k=\frac{(\boldsymbol{\mu}_k - \boldsymbol{\mu}_G)} { \|  \boldsymbol{\mu}_k - \boldsymbol{\mu}_G \|}$ and global mean $\boldsymbol{\mu}_G=\frac{1}{K}{\sum_{k=1}^{K}{\boldsymbol{\mu}_k}}$;

NC3:  The intra-class means, after centering by the global mean, align with their respective classifier weight vectors, forming the same simplex ETF. This relationship is given by $\tilde{\boldsymbol{\mu}}_k=\frac{\boldsymbol{w}_k} {\| \boldsymbol{w}_k \|}$, where $1\le k \le K$ and $\boldsymbol{w}_k$ is the classifier weight of class $k$;

NC4: When NC1-NC3 are satisfied, model predictions simplify to selecting class center closest to the feature representation. This is represented as $\text{argmax}_k\langle\mathbf{z}, \boldsymbol{w}_k\rangle=\text{argmin}_k||\mathbf{z}-\boldsymbol{\mu}_k||$, where $\langle \cdot , \cdot \rangle$ denotes the inner product operator, and $\mathbf{z}$ is the model’s output feature.

\ta{These NC attributes provide a conceptual foundation for CL. NC1-NC2 encourage compact and well-separated class representations, which help reduce interference as new classes are introduced. NC3 emphasizes the alignment between feature representations and classifier weights, promoting geometric consistency under sequential learning. As implied by NC4, such alignment induces a favorable decision geometry that simplifies classifier learning and stabilizes adaptation. Together, these properties motivate NC-based approaches for CL, which we build upon in the following section.}

\textbf{NC and Mixup.} A recent study \citep{fisher2024pushingboundariesmixupsinfluence} shows that mixup induces a distinct geometric structure. In cross-entropy classifiers with one-hot labels, same-class mixed samples align with a simplex ETF, while cross-class ones form smooth channels along decision boundaries, leading to \textbf{\textit{smoother separation}}. This effect persists even when classifier weights are fixed as a simplex ETF, enhancing both generalization and calibration.

\textbf{Inducing NC for CL}. 
Inspired by \citep{galanti_ontr_2022}, which showed that NC persists under transfer to new samples or classes, many CL studies \citep{yang2023neuralcollapse, yang2023neuralcollapseterminusunified} leverage NC via DR loss \citep{yang_induce_nc_2022} to reduce forgetting. These methods fix a simplex ETF as classifier prototypes across tasks and align them with sample features. Meanwhile, \textbf{\textit{contrastive}} learning has gained traction in CL, with remarkable performance from models like Co$^2$L \citep{co2l}, CILA \citep{wen2024provablecontrastivecontinuallearning}, and CCLIS \citep{li_cl_important_sampling_2024}. \citep{anh_wacv_2025} introduces FNC$^2$, a contrastive loss that explores NC in CL. Here, we employ both FNC$^2$ and DR losses in our analysis.

% --------------- SEC 3 - METHOD --------------
% \input{AnonymousSubmission/LaTeX/main_sec/3_method}
\section{Methodology}
\label{sec:method}

\subsection{Preliminaries}
\label{sec:preliminaries}
The general supervised CL problem is split in $T$ tasks. Each task $t \in \{1,...,T\}$ has a dataset $\mathcal{D}_t=\{(\mathbf{x}_i,y_i)\}_{i=1}^{N_t}$ with $N_t$ pairs and a distinct class set $\mathcal{C}_t$. The total number of classes across tasks: $\sum_{t=1}^{T}\left|\mathcal{C}_t\right|=K$. We focus on two popular CL settings: class-incremental learning (Class-IL) and task-incremental learning (Task-IL), where tasks have disjoint class labels: $\mathcal{C}_t \cap \mathcal{C}_{t'} = \emptyset$ for $t \neq t'$. In Task-IL, the task label is available during testing. Each batch $\mathcal{B}$ of $N$ samples is augmented to $2N$ views. After augmentation, each image $\mathbf{x}_i$ is passed through the feature extractor $f$ and projector $g$ to obtain the output $\mathbf{z}_{i}=(g \circ f)_{\theta}(\mathbf{x}_i)$ on the unit $d$-dimensional sphere (with $\theta$ denoting model parameters).

\subsection{Overall Architecture}
\label{sec:overall_architecture}
Our method is illustrated in Figure~\ref{fig:overall-methods}. We adopt a two-stage learning framework that enhances invariant representations across tasks. As in prior work \citep{co2l, wen2024provablecontrastivecontinuallearning, anh_wacv_2025}, we incorporate two main losses: (1) plasticity loss $\mathcal{L}_{plas}$ to learn new knowledge and (2) stability loss $\mathcal{L}_{stab}$ to distill past information. We predefine a set of \textit{\textbf{fixed, non-learnable}} class prototypes $\mathbf{P} = \{\mathbf{p_i}\}_{i=1}^{K}$, structured as a simplex ETF. These prototypes are shared across tasks and are used in both loss terms during training. To encourage smoother and more linear transitions between tasks, we introduce SAMix, an adaptive mixup strategy tailored for NC-based approaches. In replay settings, a memory buffer $\mathcal{M}$ is maintained using reservoir sampling \citep{reservoir_sampling_1985} to store past samples.

\ta{The overview of our method is summarized in Algorithm~\ref{alg1}. During training, each batch consists of samples from the current dataset $\mathcal{D}_t$ and the buffer $\mathcal{M}$. These are augmented and randomly selected and mixed along with their corresponding prototypes (see steps (1)-(2) in Figure~\ref{fig:overall-methods}). All views (original and mixed) are then fed into the current model $f^t$ for computing $\mathcal{L}_{plas}$ (steps (3)-(4) in Figure~\ref{fig:overall-methods}), while only original samples are passed through the frozen previous model $f^{t-1}$ to compute the stability loss $\mathcal{L}_{stab}$ (steps (5)–(6) in Figure~\ref{fig:overall-methods}). Mixed prototypes are used solely for $\mathcal{L}_{plas}$ while predefined fixed prototypes are involved in both losses.}

% ------------------ALGORITHM---------------------
\begin{algorithm}[!t]
%\small	%\textsl{}\setstretch{1.8}
    \caption{\emph{Sphere-Adaptive Mixup (SAMix) in NC-based Continual Learning.
    \ta{The key SAMix components are the generation of mixed samples and NC-based prototypes via Slerp and their selective use only in the plasticity loss.}}}
	
	% \caption{\emph{\textbf{C}ontrastive \textbf{C}ontinual \textbf{L}earning via \textbf{I}mportance \textbf{S}ampling (CCLIS).}}
	\label{alg1}
	\begin{algorithmic}[1]
        \State{\bf Input:} 
            Training sets $\{\mathcal{D}_t\}_{t=1}^T$, 
            backbone $f$, 
            projector $g$, 
            predictor $h$, 
            fixed prototypes $\mathbf{P}$, 
            hyperparameters $\upsilon$, $\iota$, $\alpha$, 
            learning rate $\eta$,
            plasticity loss type $\texttt{ploss\_type}$
		\State Initialize model $(g \circ f)_{\theta}$ and buffer $\mathcal{M}\gets \emptyset$,
            \For{$t = 1, \cdots, T$}
            \For{batch $\mathcal{B}_t\sim \mathcal{D}_t \cup \mathcal{M}$}
            % \STATE $\mathcal{B}_\mathcal{M}\sim \mathcal{M}$
		      % \STATE $\mathcal{B} \gets \mathcal{B}_t\cup \mathcal{B}_\mathcal{M}$
            \State $\lambda \gets \text{Beta}(\alpha, \alpha)$ 
            % \Comment{SAMix mixing step}
            % \Statex   \LeftComment{1} { \% first for loop service \%}
            
            \ta{\LeftComment{1}{\text{Sample and prototype mixing}}}
            \State $\mathcal{B}_{mix} \gets \text{SAMix}(\mathcal{B}_t, \lambda)$ with Eq. (\ref{eq:x_mix}) 
            \State $\mathbf{P}_{mix} \gets \text{SAMix}(\mathbf{P}, \lambda)$ with Eq. (\ref{eq:mixed_protos})
            
            \ta{\LeftComment{1}{Plasticity loss using both original and mixed samples}}
            \If{ploss\_type is \texttt{DR}}
            \State $\mathcal{L}\gets \mathcal{L}_{DR}(\theta; \mathcal{B}_t) + \upsilon \mathcal{L}_{DR}(\theta; \mathcal{B}_{mix})$ 
            
            \Else
            \State $\mathcal{L}\gets \mathcal{L}_{FNC^2}(\theta; \mathcal{B}_t) + \iota \mathcal{L}_{DR}(\theta; \mathcal{B}_{mix})$ 
            
            \EndIf
            
            \ta{\LeftComment{1}{Stability loss using only original samples}}
            
            \If{t $>$ 1}
            \State $\mathcal{L}\gets \mathcal{L} + \mathcal{L}_{HSD}(\theta; \theta_{prev}, \mathcal{B}_t)$ 
            % (see $\mathcal{L}_{HSD}$ in Eq. (~\ref{eq:hsd}))
            % (see $\mathcal{L}_{HSD}$ in the Appendix)
            \EndIf
            \State $\theta \gets \theta - \eta\nabla_{\theta} \mathcal{L}$ 
            \EndFor
            \State $\mathcal{M}\gets \text{Reservoir}(\mathcal{M}\cup \mathcal{D}_t)$
            \State $\theta_{prev}\gets \theta$
            \EndFor
		
	\end{algorithmic}  
\end{algorithm}
% ---------------------------------------

\subsection{Sphere-Adaptive Mixup}
\label{sec:phantom_method}
Our new mixup method utilizes mixed samples and prototypes to improve smooth boundaries in representation learning and enhance calibration. Each batch is drawn from the current task data $\mathcal{D}_t$ and the memory buffer $\mathcal{M}$. Given two random samples $(\mathbf{x}_i, y_i)$ and $(\mathbf{x}_j, y_j)$ with corresponding prototypes $\mathbf{p}_{y_i}$ and $\mathbf{p}_{y_j}$, we generate a mixed sample  $\tilde{\mathbf{x}}_{ij}$ as in Eq. (\ref{eq:x_mix}). Unlike original mixup \citep{mixup_2018}, which combines target labels as in Eq. (\ref{eq:y_mix}), we generate new prototypes based on the mixing coefficient $\lambda$. To generate mixed prototypes, several works \citep{yang_induce_nc_2022, fisher2024pushingboundariesmixupsinfluence} use linear interpolation, defined as:
{
\begin{equation}
    \tilde{\mathbf{p}}_{ij} = \lambda \mathbf{p}_{y_i} + (1-\lambda) \mathbf{p}_{y_j}
    \label{eq: linear-prototypes}
\end{equation}
}

% --------- FIGURE --------
\begin{figure}[t!]
  \centering
   \includegraphics[width=0.66\linewidth]{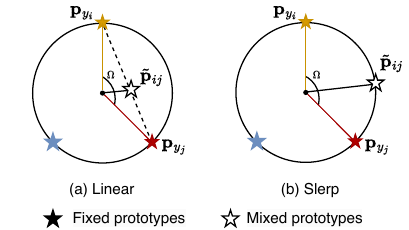}
   \caption{\centering Linear interpolation and Slerp}
   \label{fig:linear-slerp}
   % \vspace{-0.8em}
\end{figure}
% -------------------------

However, since both the features $\mathbf{z}_i, \mathbf{z}_j$ of the corresponding samples $\mathbf{x}_i, \mathbf{x}_j$ and the prototypes $\mathbf{p}_{y_i}, \mathbf{p}_{y_j}$ lie on the unit hypersphere, interpolating mixed prototypes using Eq. (\ref{eq: linear-prototypes}) is inappropriate. As illustrated in Figure~\ref{fig:linear-slerp}a, using linear interpolation causes the mixed prototype lying on the linear path between two prototypes, rather than on the unit hypersphere. The mixed prototype $\tilde{\mathbf{p}}_{ij}$ resides on the unit hypersphere only if its $l2$-norm satisfies: $\|\tilde{\mathbf{p}}_{ij}\|_2 = 1$. Based on Eq. (\ref{eq: linear-prototypes}), we compute its squared $l2$-norm as:
% {\small
\begin{align}
    \|\tilde{\mathbf{p}}_{ij}\|_2^2 &= \|\lambda \mathbf{p}_{y_i} + \left(1 - \lambda\right) \mathbf{p}_{y_j} \|_2^2 \nonumber\\
    & = \lambda^2 \|\mathbf{p}_{y_i}\|_2^2 + 2\lambda(1-\lambda) \langle\mathbf{p}_{y_i} \cdot \mathbf{p}_{y_j}\rangle + {(1 - \lambda) ^ 2}{\|\mathbf{p}_{y_j}\|_2^2}
    \label{eq:squared-l2-norm-details}
\end{align}
% }
Since $\|\mathbf{p}_{y_i}\|_2^2 = 1$ and $\|\mathbf{p}_{y_j}\|_2^2 = 1$, Eq. (\ref{eq:squared-l2-norm-details}) simplifies to:
% {\small
\begin{equation}
    \|\tilde{\mathbf{p}}_{ij}\|_2^2 = \lambda^2 + (1 - \lambda)^2 + 2{\lambda}{(1 - \lambda)}{\cos{\Omega}}
    \label{eq:squared-l2-norm-simplify}
\end{equation}
% }
    
where $\langle\cdot\rangle$ is the cosine similarity, $\Omega = \angle(\mathbf{p}_{y_i}, \mathbf{p}_{y_j})$. From Eq. (\ref{eq:squared-l2-norm-simplify}), $\tilde{\mathbf{p}}_{ij}$ only remain $l2$-normalized when $\cos{\Omega} = 1$, meaning $\mathbf{p}_i$ and $\mathbf{p}_j$ are in the same direction.  Since each mixed prototype is created from two prototypes corresponding to two randomly selected sample pairs, these two prototypes point in the same direction only if the two selected samples are from the same class. In this case, they are identical and the mixed prototype is the prototype of this class. Otherwise with two samples from different classes, the angle $\Omega \neq 0$, leading to $\|\tilde{\mathbf{p}}_{ij}\| \neq 1$, the mixed prototype is not on the unit hypersphere.

To ensure that mixed prototypes always remain on the unit hypersphere, we use Spherical Linear Interpolation (Slerp) \citep{shoemake_slerp_1985} instead of linear interpolation. A geometric visualization of this approach is shown in Figure \ref{fig:linear-slerp}b. The hypersphere-interpolated prototype $\tilde{\mathbf{p}}_{ij}$ is:
% {
\begin{equation}
    \tilde{\mathbf{p}}_{ij} = \gamma_i \mathbf{p}_{y_i} + \gamma_j \mathbf{p}_{y_j}
    \label{eq:mixed_protos}
\end{equation}
% }

where $\gamma_i = \frac{\sin (\lambda \Omega)}{\sin \Omega}$ and $\gamma_j = \frac{\sin((1-\lambda) \Omega)}{\sin \Omega}$.

Note that, due to the random selection of sample pairs, both mixed samples and mixed prototypes \textit{vary across batches}. We refer to this adaptive mixup of fixed prototypes on a hypersphere as Sphere-Adaptive Mixup (SAMix). Like the original prototypes, the mixed prototypes are \textbf{fixed and not updated} during training.

\subsection{Integration of SAMix into NC-based Methods} 
\label{sec:considerd-nc-methods}

We note that mixed samples from SAMix are used only in $\mathcal{L}_{plas}$ and \textit{not} in $\mathcal{L}_{stab}$. On the first hand, mixed samples are designed to enhance plasticity for new tasks (smoother transitions between tasks). On the other hand, $\mathcal{L}_{stab}$ focuses on preserving old knowledge: when using relation-based distillation techniques, such as instance-wise or instance-prototype relationships, only the fixed prototypes and original samples should be used, not the mixed ones. This approach aligns with the work in \citep{beyer2022knowledgedistillationgoodteacher}, which suggests that a ``patient'' teacher is ideal for knowledge distillation, meaning mixed 
% (considerd as augmented)
samples, and thus mixed class prototypes are unnecessary for distillation. \ta{Moreover, we empirically experimented with incorporating SAMix into the stability loss, but observed no performance improvement while incurring a noticeable increase in training time. Therefore, we exclude SAMix from $\mathcal{L}_{stab}$, resulting in a favorable trade-off between additional training time and performance gains, as discussed in Sec.~\ref{sec:extra-time-samix}.}

SAMix is designed to complement any NC-based method that relies on fixed prototypes. In this work, we investigate SAMix with two prominent NC-based plastic losses: dot-regression (DR) \citep{yang_induce_nc_2022} and the recent focal neural collapse contrastive (FNC$^2$) \citep{anh_wacv_2025} which are respectively defined in Eq. (\ref{eq:dr-loss}) and Eq. (\ref{eq:fnc2-loss}): 
% {\small
\begin{equation}
    \mathcal{L}_{DR} = \frac{1}{2N}\sum_{i=1}^{2N} \frac{1}{2}(\langle \mathbf{z}_i \cdot {\mathbf{p}}_{y_i}\rangle - 1)^2
    \label{eq:dr-loss}
\end{equation}
% }{\small
\begin{align}
    \mathcal{L}_{FNC^2} = -\sum_{i=1}^{2N} \frac{1}{|P(i)| + 1} \Bigg(& \sum_{\mathbf{z}_{j} \in P(i)} (1-c_{ij})^{\gamma} \log(c_{ij}) + (1-r_{i})^{\gamma} \log(r_{i}) \Bigg)
    \label{eq:fnc2-loss}
\end{align}
% }
with $c_{ij}=\frac{e^{\langle\mathbf{z}_i \cdot \mathbf{z}_j\rangle/\tau}}{A}$, $r_{i}=\frac{e^{\langle\mathbf{z}_i \cdot \mathbf{p}_{{y}_i}\rangle/\tau}}{A}$, and $A = \sum_{k \neq i}{e^{\langle\mathbf{z}_i \cdot \mathbf{z}_k\rangle/\tau}} + {\sum_{\mathbf{p}_l \in \mathbf{P}_{1:t-1}}{e^{\langle{\mathbf{z}_i \cdot \mathbf{p}_l}\rangle/\tau}}}$. Here, $P(i)$ is the set of indices for positive views (i.e., same-class views) of the anchor $\mathbf{x}_i$ in the current batch, and $\tau$ is the temperature factor; $\mathbf{p}_{{y}_i}$ is the prototype of class $y_i$, and $\mathbf{P}_{1:t-1}$ denotes the set of prototypes from past tasks.

\noindent\textbf{Training batch composition. } 
Each batch includes $2N$ normal samples (see Sec.~\ref{sec:preliminaries}) and $2N$ mixed samples ($4N$ total). Despite the increased batch size, our method achieves significant performance gains with little extra time (see Sec.~\ref{sec:extra-time-samix}) and remains effective even with smaller batches (see Sec.~\ref{sec:per-reduce-bs}).

\noindent\textbf{Mixed samples in plasticity loss.}\leavevmode\\
$\bullet$ For DR-based methods, we apply the DR loss to both normal samples ($\mathbf{z}$) and mixed samples ($\tilde{\mathbf{z}}$), yielding $\mathcal{L}_{plas} = \mathcal{L}_{DR}(\mathbf{z}) + \upsilon \mathcal{L}_{DR}(\tilde{\mathbf{z}})$, where $\upsilon$ is a trade-off factor. 

\noindent $\bullet$ For FNC$^2$-based methods, we apply $\mathcal{L}_{FNC^2}$ only to normal samples and use $\mathcal{L}_{DR}$ to mixed samples, resulting in $\mathcal{L}_{plas} = \mathcal{L}_{FNC^2}(\mathbf{z}) + \iota \mathcal{L}_{DR}(\tilde{\mathbf{z}})$, with $\iota$ as a balancing hyperparameter. This means that we do not use mixed samples as negative in contrastive loss. 

\textit{\textbf{Explanation.}} Unlike prior contrastive learning work \citep{focalcontrastive} that uses mixup to create hard negative samples, SAMix follows the original mixup approach \citep{mixup_2018}, drawing coefficients $\lambda \in [0, 1]$ from a Beta$(\alpha, \alpha)$ distribution and pairing samples randomly, regardless of class. This can produce near-positive mixed samples (e.g., when $\lambda$ is close to 0 or 1, or when pairs are from the same class), which may disrupt contrastive alignment \citep{wang2022understandingcontrastiverepresentationlearning} if treated as negatives. To prevent this, we avoid using mixed samples as negative views in losses like FNC$^2$. Moreover, applying a contrastive loss like FNC$^2$ to mixed samples is suboptimal, since each mixed sample forms a single-instance minority class, and as in the Appendix proof~\ref{sec:theoretical-model}, this approach can even disrupt the mixed-up configuration \citep{fisher2024pushingboundariesmixupsinfluence}.

\noindent $\bullet$ Mixed samples in DR loss. For the DR loss (see Eq. (\ref{eq:dr-loss})), mixed samples are directly incorporated without modification. For each of the $2N$ mixed samples in the batch, we replace the feature $\mathbf{z}_i$ and prototype $\mathbf{p}_{y_i}$ with the mixed sample feature $\tilde{\mathbf{x}}_{ij}$ and mixed prototype $\tilde{\mathbf{p}}_{ij}$, respectively, ensuring seamless integration into the loss computation.

\noindent\textbf{Stability loss.} For $\mathcal{L}_{stab}$, we use hardness-softness distillation (HSD) \citep{anh_wacv_2025}, which surpasses instance-wise relation distillation \citep{co2l} and sample-prototype relation distillation (S-PRD) \citep{anh_wacv_2025} in preserving old knowledge (see Appendix~\ref{sec:def-distillation-losses} for loss definitions). We refer to the approach in \citep{anh_wacv_2025} as Focal Contrastive Neural Collapse Continual Learning (FC-NCCL). Besides, we propose a novel method using DR for plasticity and HSD for stability, which we refer to Tightly Aligned Neural Collapse Continual Learning (TA-NCCL).
% \vspace{-0.28em}
\subsection{Expected Outcome Hypothesis}
\label{sec:expected-outcome-hypothesis}
We hypothesize the following results, to be validated in Sec. \ref{sec:experiments}, confirming SAMix's role in improving CL performance and calibration, and in reducing overconfidence.

\textbf{\textit{(a) H1: Performance without SAMix.}} FC-NCCL is expected to outperform TA-NCCL, as it better preserves within-class distributions during representation learning \citep{anh_wacv_2025}. It is also anticipated to outperform TA-NCCL in calibration, since its FNC$^2$ loss already integrates temperature scaling and focal loss, two well-established calibration techniques \citep{guo2017_calibration_nn_2017, lin2018focallossdenseobject}.

\textbf{\textit{(b) H2: Limited benefit of SAMix for FNC$^2$.}}
We hypothesize that SAMix provides limited benefits for FNC$^2$ in FC-NCCL when FNC$^2$ is applied to normal samples and DR to mixed ones. As a contrastive loss, FNC$^2$ benefits from more negative samples \citep{simclr-chen20j}, which SAMix does not provide. Therefore, we predict that any performance gains from integrating SAMix primarily come from the DR loss, as it better leverages mixed samples. In contrast, FNC$^2$ relies on a large number of negative samples, which SAMix does not address.

\textbf{\textit{(c) H3: Significant benefit of SAMix for DR.}}
In contrast, we expect DR, which operates on mixed samples, to fully benefit from SAMix. Unlike FNC$^2$, which considers both sample-sample and sample-prototype contrastive relationships for flexible clustering, DR only focuses on sample-prototype relations, resulting in distinct but less flexible clusters. However, when SAMix is applied, the decision boundaries become smoother. As mentioned in \citep{fisher2024pushingboundariesmixupsinfluence}, using mixup can create new configurations with mixed samples, significantly enhancing representation learning. We expect this will allow DR to learn more robust representations, improve generalization, and perform better on both the current task and across tasks in CL, while also mitigating forgetting.

\textbf{\textit{(d) H4: Enhanced calibration and reduced overconfidence.}}
We further hypothesize that SAMix improve calibration and reduce overconfidence for both FC-NCCL and TA-NCCL, at least on the current task. This stems from the inherent calibration properties of mixup, which SAMix inherits. SAMix's ability to smooth decision boundaries and create mixed samples will lead to more calibrated predictions, helping to reduce overconfidence.

% --------------- SEC 4 - EXPERIMENTS --------------
% \input{AnonymousSubmission/LaTeX/main_sec/4_experiments}
\section{Experiments}
\label{sec:experiments}
% --------------- RESULT TABLE -----------------
\begin{table*}[!t]
    \caption{Comparison of our method with supervised baselines at memory sizes 200 and 500, averaged over 5 trials from 5 different seeds (best results in bold). $\blacktriangle$ indicates performance improvement when applying SAMix. 
    % DR loss is used for all SAMix-generated samples. 
    (\textbf{\text{Ours: TA-NCCL}} method)}
    \centering
    % \resizebox*{!}{1.0\columnwidth}{
    % \input{AnonymousSubmission/LaTeX/tables/table_acc_result}
    % }
    \small
    \setlength{\tabcolsep}{1.2mm}  %1.6pt \renewcommand{\arraystretch}{1.2}
    \renewcommand{\arraystretch}{1.1}
    \resizebox{1.0\linewidth}{!}{
    \begin{tabular}{clccccccc}
    \hline
    
    \multirow{2}{*}{\textbf{Buffer}} & \textbf{Dataset} & \multicolumn{2}{c}{\textbf{Seq-Cifar-100}} & \multicolumn{2}{c}{\textbf{Seq-Tiny-ImageNet}} & \multicolumn{2}{c}{\textbf{Seq-Cifar-10}}\\
                 & \textbf{Scenario} & \textbf{Class-IL} & \textbf{Task-IL} & \textbf{Class-IL} & \textbf{Task-IL} & \textbf{Class-IL} & \textbf{Task-IL}\\
    \hline
        \multirow{7}{*}{0}
        & Co$^2$L (ICCV'21) \citepp{co2l} &  26.89$\pm$0.78   &  51.91$\pm$0.63  &  13.43$\pm$0.57  &  40.21$\pm$0.68  &  58.89$\pm$2.61  &  86.65$\pm$1.05\\
        \cmidrule{2-8}
        & \grayrow{FC-NCCL (WACV'25) \citepp{anh_wacv_2025}} {32.57$\pm$0.55} {57.87$\pm$0.62} {14.54$\pm$0.52} {43.81$\pm$0.47} {69.26$\pm$0.32} {94.41$\pm$0.43}\\
    
        & \grayrow{FC-NCCL + SAMix} {40.15$\pm$0.48} {64.96$\pm$0.72} {15.72$\pm$0.34} {44.65$\pm$0.59} {66.47$\pm$1.26} {93.09$\pm$1.13}\\
    
        & \grayrow {} {\text{(+7.58 $\blacktriangle$)}} {\text{(+7.09 $\blacktriangle$)}} {\text{(+1.18 $\blacktriangle$)}} {\text{(+0.84 $\blacktriangle$)}} {\text{(-2.79)}} {\text{(-1.32)}} \\
    
        & \grayrow {Ours} {33.48$\pm$0.22} {60.79$\pm$0.54} {10.88$\pm$0.51} {35.61$\pm$0.41} {64.83$\pm$0.56} {94.54$\pm$0.48}\\
    
        & \grayrow {Ours + SAMix} {\textbf{44.41$\pm$0.35}} {\textbf{70.33$\pm$0.48}} {\textbf{17.19$\pm$0.48}} {\textbf{45.12$\pm$0.73}} {\textbf{70.32$\pm$0.43}} {\textbf{95.89$\pm$0.74}}\\
    
        & \grayrow {} {\text{(+10.93 $\blacktriangle$)}} {\text{(+9.54 $\blacktriangle$)}} {\text{(+6.31 $\blacktriangle$)}} {\text{(+9.51 $\blacktriangle$)}} {\text{(+5.49 $\blacktriangle$)}} {\text{(+1.35 $\blacktriangle$)}} \\
    
    \hline
        \multirow{12}{*}{200}
        & ER (ICLR'19) \citepp{riemer2019learninglearnforgettingmaximizing} & 21.78$\pm$0.48 & 60.19$\pm$1.01 & 8.49$\pm$0.16 & 38.17$\pm$2.00 & 44.79$\pm$1.86 & 91.19$\pm$0.94\\
    
        % & iCaRL \citep{rebuffi2017icarl} & CVPR'17 & 28.00$\pm$0.91 & 51.43$\pm$1.47 & 7.53$\pm$0.79 & 28.19$\pm$1.47 & 49.02$\pm$3.20 & 88.99$\pm$2.13\\
    
        % & GEM \citep{lopezpaz2017gem} & 20.75$\pm$0.66 & 58.84$\pm$1.00 & - & - & 25.54$\pm$0.76 & 90.44$\pm$0.94\\
    
        % & GSS \citep{aljundi2019gradientbasedsampleselection} & 19.42$\pm$0.29 & 55.38$\pm$1.34 & - & - & 39.07$\pm$5.59 & 88.80$\pm$2.89\\
    
        & DER (NeurIPS'20) \citepp{buzzega2020darkexperiencegeneralcontinual} & 31.23$\pm$1.38 & 63.09$\pm$1.09 & 11.87$\pm$0.78 & 40.22$\pm$0.67 & 61.93$\pm$1.79 & 91.40$\pm$0.92\\
    
        & Co$^2$L (ICCV'21) \citepp{co2l} & 27.38$\pm$0.85 & 53.94$\pm$0.76 & 13.88$\pm$0.40 & 42.37$\pm$0.74 & 65.57$\pm$1.37 & 93.43$\pm$0.78\\
    
        & GCR (CVPR'22) \citepp{tiwari2022gcrgradientcoresetbased} & 33.69$\pm$1.40 & {64.24$\pm$0.83} & 13.05$\pm$0.91 & 42.11$\pm$1.01 & 64.84$\pm$1.63 & 90.80$\pm$1.05\\
    
        & CILA (ICML'24) \citepp{wen2024provablecontrastivecontinuallearning} & - & - & 14.55$\pm$0.39 & 44.15$\pm$0.70 & 67.06$\pm$1.59 & 94.29$\pm$0.24\\
    
        & {CCLIS} (AAAI'24) \citepp{li_cl_important_sampling_2024} & {42.39$\pm$0.37} & {72.93$\pm$0.46} & {16.13$\pm$0.19} & {48.29$\pm$0.78} & \textbf{74.95$\pm$0.61} & {96.20$\pm$0.26}\\ 
    
        \cmidrule{2-8}
        &\grayrow {FC-NCCL (WACV'25) \citepp{anh_wacv_2025}} {34.04$\pm$0.42} {59.46$\pm$0.65} {15.52$\pm$0.53} {44.59$\pm$0.72} {72.63$\pm$0.78} {95.31$\pm$0.32}\\
    
        &\grayrow  {FC-NCCL + SAMix} {41.92$\pm$0.68} {66.36$\pm$0.70} {17.58$\pm$0.35} {46.83$\pm$0.59} {69.31$\pm$0.68} {94.93$\pm$0.74}\\
    
        &\grayrow {} {\text{(+7.88 $\blacktriangle$)}} {\text{(+6.90 $\blacktriangle$)}} {\text{(+2.06 $\blacktriangle$)}} {\text{(+2.24 $\blacktriangle$)}} {\text{(-3.32)}} {\text{(-0.38)}}\\
    
        &\grayrow  {Ours} {36.67$\pm$1.18} {64.64$\pm$1.37} {11.04$\pm$0.29} {35.49$\pm$0.62} {66.78$\pm$0.23} {93.98$\pm$0.17}\\
    
        &\grayrow  {Ours + SAMix} {\textbf{46.95$\pm$0.71}} {\textbf{73.48$\pm$0.29}} {\textbf{18.39$\pm$0.38}} {\textbf{49.51$\pm$0.83}} {73.25$\pm$0.54} {\textbf{96.68$\pm$0.32}}\\
    
        &\grayrow  {} {\text{(+10.28 $\blacktriangle$)}} {\text{(+8.84 $\blacktriangle$)}} {\text{(+7.35 $\blacktriangle$)}} {\text{(+14.02 $\blacktriangle$)}} {\text{(+6.47 $\blacktriangle$)}} {\text{(+2.70 $\blacktriangle$)}}\\
    \hline
        \multirow{12}{*}{500}
        & ER (ICLR'19) \citepp{riemer2019learninglearnforgettingmaximizing} & 27.66$\pm$0.61 & 66.23$\pm$1.52 & 9.99$\pm$0.29 & 48.64$\pm$0.46 & 57.74$\pm$0.27 & 93.61$\pm$0.27\\
    
        % & iCaRL \citep{rebuffi2017icarl} & CVPR'17 & 33.25$\pm$1.25 & 58.16$\pm$1.76 & 9.38$\pm$1.53 & 31.55$\pm$3.27 & 47.55$\pm$3.95 & 88.22$\pm$2.62\\
    
        % & GEM \citep{lopezpaz2017gem} & 25.54$\pm$0.65 & 66.31$\pm$0.86 & - & - & 26.20$\pm$1.26 & 92.16$\pm$0.64\\
    
        % & GSS \citep{aljundi2019gradientbasedsampleselection} & 21.92$\pm$0.34 & 60.28$\pm$1.18 & - & - & 49.73$\pm$4.78 & 91.02$\pm$1.57\\
    
        & DER (NeurIPS'20) \citepp{buzzega2020darkexperiencegeneralcontinual} & 41.36$\pm$1.76 & 71.73$\pm$0.74 & 17.75$\pm$1.14 & 51.78$\pm$0.88 & 70.51$\pm$1.67 & 93.40$\pm$0.39\\
    
        & Co$^2$L (ICCV'21) \citepp{co2l} & 37.02$\pm$0.76 & 62.44$\pm$0.36 & 20.12$\pm$0.42 & 53.04$\pm$0.69 & 74.26$\pm$0.77 & 95.90$\pm$0.26\\
    
        & GCR (CVPR'22) \citepp{tiwari2022gcrgradientcoresetbased} & 45.91$\pm$1.30 & {71.64$\pm$2.10} & 19.66$\pm$0.68 & 52.99$\pm$0.89 & 74.69$\pm$0.85 & 94.44$\pm$0.32\\

        & CILA (ICML'24) \citepp{wen2024provablecontrastivecontinuallearning} & - & - & 20.64$\pm$0.59 & {54.13$\pm$0.72} & {76.03$\pm$0.79} & {96.40$\pm$0.21}\\
    
        & {CCLIS} (AAAI'24) \citepp{li_cl_important_sampling_2024} & {46.08$\pm$0.67} & {74.51$\pm$0.38} & {22.88$\pm$0.40} & {57.04$\pm$0.43} & {78.57$\pm$0.25} & {96.18$\pm$0.43}\\ 
    
        \cmidrule{2-8}
        &\grayrow  {FC-NCCL (WACV'25) \citepp{anh_wacv_2025}} {{40.25$\pm$0.58}} {65.85$\pm$0.44} {{20.31$\pm$0.34}} {{53.46$\pm$0.59}} {75.51$\pm$0.52} {96.14$\pm$0.25}\\
    
        &\grayrow  {FC-NCCL + SAMix} {{47.87$\pm$0.57}} {72.95$\pm$1.05} {20.71$\pm$0.43} {54.25$\pm$0.57} {75.87$\pm$0.66} {95.35$\pm$0.39}\\
        
        & \grayrow {} {\text{(+7.62 $\blacktriangle$)}} {\text{(+7.10 $\blacktriangle$)}} {\text{(+0.40 $\blacktriangle$)}} {\text{(+0.79 $\blacktriangle$)}} {\text{(+0.36 $\blacktriangle$)}} {\text{(+0.79 $\blacktriangle$)}} \\
        
        &\grayrow  {Ours} {41.72$\pm$0.16} {68.42$\pm$0.33} {16.05$\pm$0.36} {44.22$\pm$0.89} {68.71$\pm$0.59} {94.68$\pm$0.52}\\
    
        &\grayrow  {Ours + SAMix} {\textbf{55.85$\pm$0.61}} {\textbf{79.02$\pm$0.78}} {\textbf{22.95$\pm$0.47}} {\textbf{57.13$\pm$0.32}} {\textbf{78.62$\pm$0.38}} {\textbf{96.98$\pm$0.57}}\\
        
        & \grayrow {} {\text{(+14.13 $\blacktriangle$)}} {\text{(+10.60 $\blacktriangle$)}} {\text{(+6.90 $\blacktriangle$)}} {\text{(+12.91 $\blacktriangle$)}} {\text{(+9.91 $\blacktriangle$)}} {\text{(+2.30 $\blacktriangle$)}} \\
    \hline
    \end{tabular}
    }

    \label{tab:acc_result}
\end{table*}
% ----------------------------------------------

% -------------- FIGURE AOE-AECE-Acc RESULTS ----------
\begin{figure*}[t!]
    \centering
    \includegraphics[width=1.0\linewidth]{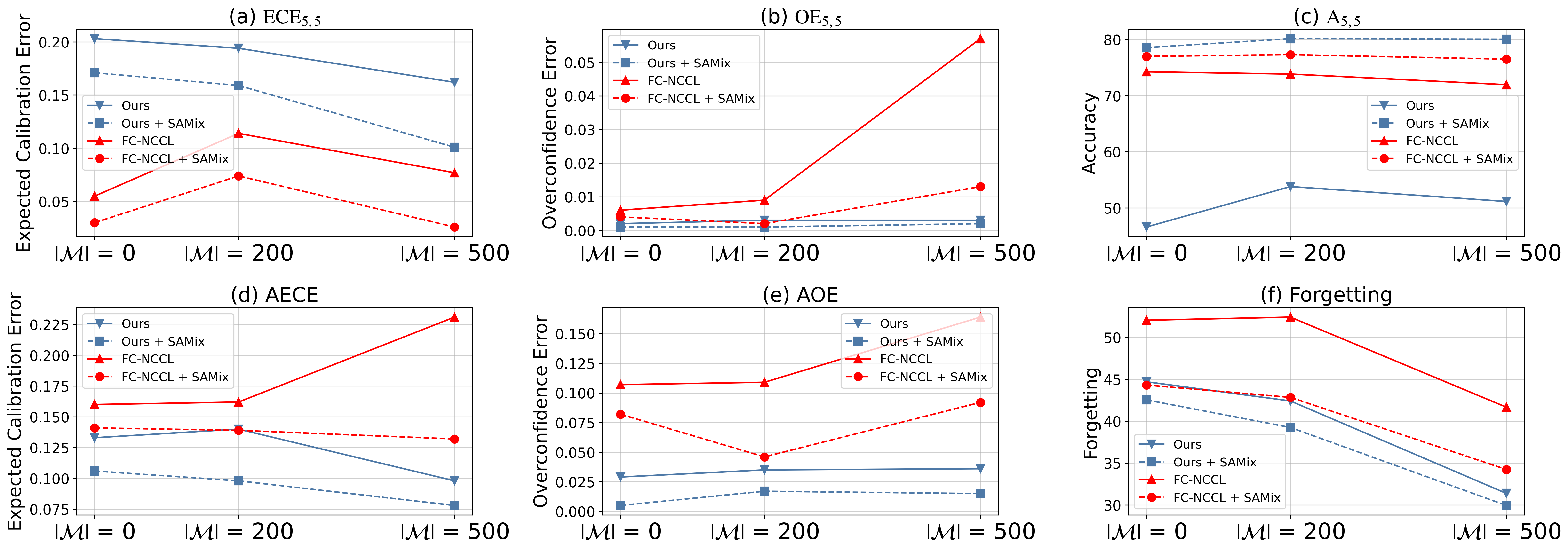}
    % \captionsetup{justification=centering}
    \caption{\centering Results on Seq-Cifar100 with the Class-IL setting}
    \label{fig:calibration-results}
   % \vspace{-0.8em}
\end{figure*}
% ----------------------------------------------------

\subsection{Datasets and Implementation details}
\noindent \textbf{Datasets.} We use three standard CL datasets widely used in literature for fair comparison under Class-IL and Task-IL settings (\ref{sec:preliminaries}): Seq-Cifar-10 and Seq-Cifar-100 \citep{krizhevsky2009learning}, and Seq-Tiny-ImageNet \citep{le2015tinyiv}. Seq-Cifar-10 has 5 tasks with 2 classes each; Seq-Cifar-100 and Seq-Tiny-ImageNet each have 5 and 10 tasks with 20 classes per task, respectively.

\noindent \textbf{Implementation details.} We utilize ResNet-18 \citep{resnet18} as the backbone, omitting the final layer as in \citep{co2l, fini2022cassle, anh_wacv_2025, li_cl_important_sampling_2024}. Following \citep{simclr-chen20j, zbontar2021barlowtwinsselfsupervisedlearning, co2l}, we add a two-layer projection MLP $g^t$ to map backbone features into a $d$-dimensional space ($d=128$ for Seq-Cifar-10/100, $d=256$ for Seq-Tiny-ImageNet). For distillation, as in \citep{fini2022cassle}, we use a predictor $h^t$ with the same architecture as $g^t$ to align current and past feature spaces. Besides, we use buffer sizes of 0, 200, and 500. For evaluation, following \citep{co2l, wen2024provablecontrastivecontinuallearning, anh_wacv_2025}, we train a classifier $s^t$ on top of the frozen encoder $f^t$ for 100 epochs, using samples from the last-task training dataset $\mathcal{D}_t$ and buffer $\mathcal{M}$. As in \citep{co2l}, no buffer is used during representation learning in the memory-free setting, but 200 auxiliary samples are still needed to only train the classifier for all datasets. Experiments are conducted on a NVIDIA Quadro GV100 GPU with 256GB RAM, Ubuntu 20.04, and PyTorch 1.12.1 \citep{paszke2019pytorchimperativestylehighperformance}.

\noindent\textbf{Considered methods.} We evaluate SAMix across the SOTA NC-based methods: FC-NCCL \citep{anh_wacv_2025} and TA-NCCL. We also consider recent baselines, including CCLIS \citep{li_cl_important_sampling_2024}, CILA \citep{wen2024provablecontrastivecontinuallearning}, and other well-known supervised methods such as ER \citep{riemer2019learninglearnforgettingmaximizing}, 
% iCaRL \citep{rebuffi2017icarl}, 
% GEM \citep{lopezpaz2017gem}, GSS \citep{aljundi2019gradientbasedsampleselection}, 
DER \citep{buzzega2020darkexperiencegeneralcontinual}, Co$^2$L \citep{co2l}, and GCR \citep{tiwari2022gcrgradientcoresetbased}.

%-------------------------------------------------------------------------
\noindent \textbf{Evaluation metrics.} 
As in prior work \citep{co2l, fini2022cassle, wen2024provablecontrastivecontinuallearning, anh_wacv_2025, li_cl_important_sampling_2024}, we report Average Accuracy (AA) \citep{chaudhry_2018} across all tasks on the test dataset after learning the final task $T$: $\text{AA} = \frac{1}{T}\sum_{k=1}^{T} \text{A}_{T,k}$. In addition, we use Average Forgetting \citep{chaudhry_2018} to measure how much the model forgets previous tasks (see Appendix~\ref{sec:average-forgetting}).

\noindent\textbf{Calibration metrics.} To assess network calibration, we use the metrics from \citep{guo2017_calibration_nn_2017}. Predictions are divided into $M$ interval equal-sized bins, with $B_m$ representing the set of samples whose prediction scores (the winning softmax score) fall into bin $m$. The accuracy and confidence of $B_m$ are defined as: $\text{acc}(B_m) = \frac{1}{|B_m|}\sum_{i \in B_m} \mathbf{1}(\hat{y}_i = y_i)$; $\text{conf}(B_m) = \frac{1}{|B_m|}\sum_{i \in B_m} \hat{p}_i$,
where $\hat{p}_i$ is the confidence (winning score) of sample $i$. The {Expected Calibration Error} (ECE) is defined as: $\text{ECE} = \sum_{m=1}^M \frac{|B_m|}{n}  | \text{acc}(B_m) - \text{conf}(B_m) |$. We also use the Overconfidence Error (OE) metric \citep{thulasidasan_on_mixup_training_2020}, essential for high-risk applications where confidently incorrect predictions are critical: $\text{OE} = \sum_{m=1}^M \frac{|B_m|}{n} [\text{conf}(B_m) \times \max ( \text{conf}(B_m) - \text{acc}(B_m), 0 ) ]$. This metric heavily penalizes overconfident predictions when confidence exceeds accuracy. 

Building on these metrics, we propose new metrics which are adaptive to CL: {\bf Average Expected Calibration Error} (AECE) and {\bf Average Overconfidence Error} (AOE):
{\small
\begin{equation}
    \text{AECE} = \frac{1}{T} \sum_{k=1}^{T}{\text{ECE}_{T, k}};
    \text{AOE} = \frac{1}{T} \sum_{k=1}^{T}{\text{OE}_{T, k}}
\end{equation}
}
where $\text{ECE}_{T,k}$, $\text{OE}_{T,k}$ is the ECE, OE value of task $k$ after learning the final task $T$ correspondingly.

% ------------------------ SEC: MAIN RESULTS ----------------------
\subsection{Main Results}

Experimental results are shown in Table \ref{tab:acc_result}, where we report accuracies from baselines with published or available code. Figure 3 presents ECE$_{5,5}$, OE$_{5,5}$, and A$_{5,5}$ on the final task ($T = 5$) (top row), and AECE, AOE, Forgetting across tasks (bottom row) on Seq-Cifar-100. These results highlight the impact of SAMix on FNC$^2$ and DR, as well as on calibration and performance. We evaluate these findings against the proposed hypotheses, focusing on FC-NCCL and TA-NCCL.

\textbf{a) H1 - without SAMix.} The hypothesis is \textbf{partially} validated.
On Seq-Tiny-ImageNet and Seq-Cifar-10, FC-NCCL outperforms TA-NCCL in accuracy and calibration across buffer sizes, except in a few Task-IL cases where the results are nearly equivalent. This support the hypothesis that preserving within-class distributions improves CL performance. The integration of temperature scaling and focal loss in FNC$^2$ yields better-calibrated predictions, as evidenced by FC-NCCL’s significantly lower ECE$_{5,5}$ compared to DR.

However, on Seq-Cifar-100, the trend reverses, with DR outperforming FNC$^2$ entirely, which contradicts the initial performance hypothesis. We attribute this to Cifar-100’s higher inter-class similarity \citep{bertinetto2018metalearning}, stemming from its 100 classes grouped into 20 semantically related superclasses (e.g., vehicles, animals). In contrast, Tiny ImageNet contains 200 more diverse and ungrouped classes. This similarity in Cifar-100 may lead to greater class confusion, limiting the advantages of FC-NCCL. Moreover, as shown in Figure \ref{fig:calibration-results}a, FNC$^2$ achieves lower ECE$_{5,5}$ on the current task, indicating better calibration than DR. However, with buffer usage, FC-NCCL shows higher OE$_{5,5}$ than TA-NCCL (see Figure \ref{fig:calibration-results}b).

\textbf{b) H2: Limited benefit of SAMix for FNC2.} 
This hypothesis holds partially, as SAMix yields slight improvements on Seq-Tiny-ImageNet and even degrades performance on Seq-Cifar-10 with buffer sizes of 0 and 200. The degradation on Seq-Cifar-10 is mainly due to its setting, where each task contains only two classes. In memory-free or low-memory scenarios, mixed samples thus lack diversity, as they primarily come from these two classes. Furthermore, the number of mixed samples per batch equals the batch size, making it challenging for FNC$^2$, which relies on the ``push and pull'' mechanism, to achieve the \textit{alignment} and \textit{uniformity} properties \citep{wang2022understandingcontrastiverepresentationlearning} required for effective contrastive representation learning. As a result, FNC$^2$ struggles to pull same-class samples together and learn distinct, cluster-based representations. 

Moreover, in methods using plasticity loss FNC$^2$, applying DR to mixed samples often shift the model’s focus toward blended artifacts rather than meaningful features, leading to forgetting. In contrast, on Seq-Cifar-100, SAMix significantly enhances FC-NCCL performance, which is attributed to the dataset’s high inter-class similarity \citep{bertinetto2018metalearning}. In this case, SAMix refines class boundaries and boosts performance. Without it, performance drops substantially compared to DR-based, which further highlights its benefits.

\textbf{c) H3: Significant benefit of SAMix for DR.} 
Experimental results strongly support this hypothesis. With SAMix, TA-NCCL (relying on DR)  outperforms FC-NCCL, whereas without SAMix, FC-NCCL performs comparably or slightly better. These results align with our analysis in Sec. \ref{sec:expected-outcome-hypothesis}, where DR benefits more from SAMix than FNC$^2$ because DR loss strictly pulls samples toward prototypes as optimal points, leading to better representation learning, smoother decision boundaries, and improved accuracy.

\textbf{Notably, TA-NCCL with SAMix achieves SOTA results} in most settings, except on Seq-Cifar-10 (Class-IL, buffer size 200), particularly on large, complex datasets with many classes as Seq-Cifar-100 and Seq-Tiny-ImageNet. 

\textbf{d) H4: Enhanced calibration and reduced overconfidence.}
Figure \ref{fig:calibration-results} shows that SAMix significantly reduces calibration error and overconfidence while improving performance on both the current task and across tasks, effectively mitigating forgetting for both FC-NCCL and TA-NCCL in most cases. 

\textbf{Mixup only within the current task.} Without memory, SAMix applies only to current-task classes but still effectively reduces calibration error and overconfidence error while boosting performance. This highlights SAMix’s potential for CL tasks with limited storage or strict data privacy constraints.

% 2
\subsection{Ablation Studies}
\label{sec:ablation-studies}
% -------------- FIGURE ABLATION MAIN ----------
\begin{table*}
    \caption{Comparison between Slerp and linear interpolation on Seq-Cifar-100. \text{$\blacktriangle$} indicates that Slerp performs better}
    \centering
    \small
  % {\small{
  % }}
    % \setlength{\tabcolsep}{1mm}
    % \resizebox*{!}{0.41\columnwidth}{
    %     \input{AnonymousSubmission/LaTeX/tables/table_ablation_linear_slerp}
    % }
    % \resizebox{0.92\columnwidth}{!}{
    \begin{tabularx}{0.9\textwidth} {Y Y Y Y Y Y}
        \toprule
        \multirow{3}{*}{\textbf{Buffer}} & \multirow{3}{*}{\textbf{Metrics}} & \multicolumn{4}{c}{\textbf{Method}} \\
        \cmidrule{3-6}
         &  & \multicolumn{2}{c}{\textbf{FC-NCCL + SAMix}} & \multicolumn{2}{c}{\textbf{Ours + SAMix}} \\
        \cmidrule(r){3-4} \cmidrule(l){5-6}
         &  & \textbf{Linear} & \textbf{Slerp} & \textbf{Linear} & \textbf{Slerp} \\
        \midrule
        \multirow{3}{*}{0} & AECE($\downarrow$) & 0.164 & 0.141 \text{$\blacktriangle$} & 0.130 & 0.106 \text{$\blacktriangle$}\\
         & AOE($\downarrow$) & 0.108 & 0.082 \text{$\blacktriangle$} & 0.027 & 0.005 \text{$\blacktriangle$}\\
         & AA($\uparrow$) & 38.39 & 40.15 \text{$\blacktriangle$} & 42.36 & 44.41 \text{$\blacktriangle$}\\
        \midrule
        \multirow{3}{*}{200} & AECE($\downarrow$) & 0.140 & 0.139 \text{$\blacktriangle$} & 0.114 & 0.098 \text{$\blacktriangle$}\\
         & AOE($\downarrow$) & 0.071 & 0.046 \text{$\blacktriangle$} & 0.043 & 0.017 \text{$\blacktriangle$}\\
         & AA($\uparrow$) & 39.03 & 41.92 \text{$\blacktriangle$} & 43.76 & 46.95 \text{$\blacktriangle$}\\
        \midrule
        \multirow{3}{*}{500} & AECE($\downarrow$) & 0.158 & 0.132 \text{$\blacktriangle$} & 0.079 & 0.078 \text{$\blacktriangle$}\\
         & AOE($\downarrow$) & 0.105 & 0.092 \text{$\blacktriangle$} & 0.042 & 0.015 \text{$\blacktriangle$}\\
         & AA($\uparrow$) & 45.94 & 47.87 \text{$\blacktriangle$} & 53.96 & 55.85 \text{$\blacktriangle$}\\
        \bottomrule
    \end{tabularx}
    % }

    \label{tab:ablation-linear-slerp}
    % \vspace{-0.7em}
\end{table*}
% --------------
\subsubsection{Effectiveness of using slerp in SAMix}
% \noindent\textbf{Effectiveness of using slerp in SAMix.} 
We report ablation results in Table \ref{tab:ablation-linear-slerp}, comparing Slerp and linear interpolation when applying SAMix to both FC-NCCL and TA-NCCL on Seq-Cifar-100 with buffer sizes of 0, 200, and 500. Slerp yields better calibration and lower AOE in all cases than linear interpolation. Moreover, it also consistently improves accuracy at larger distances. These results support the use of Slerp in SAMix and align with our analysis in Sec.~\ref{sec:phantom_method}.

\subsubsection{Extra time when using SAMix}
\label{sec:extra-time-samix}
Although our method increase batch size (see Sec.~\ref{sec:considerd-nc-methods}), mixed samples are used only in the plasticity loss (DR), not in distillation or contrastive losses like FNC$^2$ to avoid disrupting the mixed-up configuration (see Appendix proof~\ref{sec:theoretical-model}). Thus, mixed samples are not used as regular inputs, keeping the impact on training time minimal. Table \ref{tab:ablation-extra-time} shows the extra time when applying SAMix to FC-NCCL and TA-NCCL on Seq-Cifar-100 (CIL setting). Across buffer sizes, the additional time is a good trade-off, justified by significant performance gains. As in the discussion of \citep{chen2025adaptiveretentioncorrection}, other SOTA CL methods such as L2P \citep{wang2022learningpromptcontinuallearning} and DualPrompt \citep{wang2022dualpromptcomplementarypromptingrehearsalfree} increase training time by $37\%$ and $32\%$, respectively, for performance gains of only $2.5\%$ and $1.4\%$.

\subsubsection{Performance with reduced batch size}
\label{sec:per-reduce-bs}
% \noindent\textbf{Performance with reduced batch size.} 
To evaluate SAMix under smaller batches, we run TA-NCCL on Seq-Cifar-10 (Class-IL setting) with a batch size of 256 with different buffer sizes. As shown in Table \ref{tab:ablation-reduce-bs}, even \textbf{without} memory, SAMix with half the batch size outperforms the SOTA method CCLIS \citep{li_cl_important_sampling_2024}, which uses a batch size of 512 and memory size of 200. This highlights the strong adaptability of our method under constrained settings.

\begingroup
\hfuzz=10pt
\begin{table*}[ht]
    \caption{Extra time (overhead, $\%$) and performance gain ($\%$) \textbf{\textit{with SAMix}} on Seq-Cifar-100 (CIL setting)}
    \centering
    % \setlength{\tabcolsep}{0.6mm}
    % \renewcommand{\arraystretch}{1.1}
    % \resizebox{0.5\linewidth}{!}{
    \small
        \begin{tabularx}{0.9\textwidth}{l Y Y Y Y Y Y}
            \toprule
            % \multirow{2}{*}{\diagbox{\textbf{Method}}{\textbf{Buffer}}}
             \textbf{Buffer} & \multicolumn{2}{c}{\textbf{0}} & \multicolumn{2}{c}{\textbf{200}} & \multicolumn{2}{c}{\textbf{500}} \\
            % \cline{2-7}
            \cmidrule(r){2-3} \cmidrule(r){4-5} \cmidrule{6-7}
             & \textbf{Overhead} & \textbf{Gain} & \textbf{Overhead} & \textbf{Gain} & \textbf{Overhead} & \textbf{Gain} \\
            \midrule
            FC-NCCL & 14.1 & 7.6 & 16.5 & 7.9 & 19.8 & 7.6 \\
            \midrule
            TA-NCCL & 22.3 & 10.9 & 21.7 & 10.2 & 28.1 & 14.1 \\
            \bottomrule
        \end{tabularx}
    % }
    \label{tab:ablation-extra-time}
\end{table*}
\endgroup

\begin{table}[ht]
    \caption{SAMix results with varying batch sizes on Seq-Cifar-100 (CIL setting). Bold indicates comparison purposes}
    \centering
    \setlength{\tabcolsep}{6mm}
    \small
    % \resizebox*{!}{0.3\columnwidth}{
    %     \input{AnonymousSubmission/LaTeX/tables/table_reduce_bsz}
    % }
    \begin{tabular}{cccc}
        \toprule
        \textbf{Buffer} & \textbf{Method} & \textbf{Batch size} & \textbf{AA} \\
        \midrule
        \multirow{2}{*}{0} 
        % & Co$^2$L & 512 &  26.89\\
         & TA-NCCL & 512 &  44.41\\
         & TA-NCCL & 256 & \textbf{42.77}\\
        \midrule
        \multirow{3}{*}{200} & CCLIS & 512 & \textbf{42.39}\\
         & TA-NCCL & 512 & 46.95\\
        & TA-NCCL & 256 & 45.21\\
        \bottomrule
    \end{tabular}
    \label{tab:ablation-reduce-bs}
\end{table}

% --------------- SEC 5 - CONCLUSION --------------
% \input{AnonymousSubmission/LaTeX/main_sec/5_conclusion}
\section{Conclusion}
\label{sec:conclusion}

In this work, we explore the role of mixup in enhancing model calibration, reducing forgetting, and improving performance in Neural Collapse-based continual learning. We propose SAMix, an adaptive mixup strategy tailored for NC-based CL methods, and 
% demonstrate its substantial benefits 
show its substantial benefits when paired with DR loss. SAMix significantly boosts CL model performance, achieving SOTA results on benchmark datasets such as Seq-Cifar-100 and Seq-Tiny-ImageNet. Moreover, it reduces calibration error and overconfidence across tasks, even in memory-free settings, making it particularly valuable under strict data privacy constraints.

However, a limitation exists: while SAMix improves performance and calibration, it is tailored to fixed prototype-based CL. Our experiments with dynamic prototype methods and traditional one-stage CL approaches reveal integration challenges: direct incorporation is either technically infeasible (e.g., in one-stage supervised CL) or degrades performance, as mixup on randomly initialized dynamic prototypes lacks meaningful impact, especially early in training. Although NC-based methods currently lead the field, this constraint highlights the need to extend SAMix to dynamic prototype-based methods like CCLIS \citep{li_cl_important_sampling_2024}. \ta{Moreover, like other NC-based CL methods, SAMix assumes prior knowledge of the total number of classes to predefine prototypes. Recent work \citep{markou2024guidingneuralcollapseoptimising} on dynamically aligning classifiers to the nearest simplex ETF suggests that such structures need not be fixed in advance. Extending SAMix to incrementally grow the simplex as new classes emerge in CL is a promising direction for future work.}

\backmatter

% \bmhead{Supplementary information}

% If your article has accompanying supplementary file/s please state so here. 

% Authors reporting data from electrophoretic gels and blots should supply the full unprocessed scans for key as part of their Supplementary information. This may be requested by the editorial team/s if it is missing.

% Please refer to Journal-level guidance for any specific requirements.

\bmhead{Acknowledgements}
This work is partially supported by the ANR-21-ASRO-0003 ROV-Chasseur project.

\begin{appendices}

\section{Theoretical Proof}
\label{sec:theoretical-model}

% \subsection{FNC$^2$ and DR loss}
In this section, we prove the theory that DR loss should be used instead of contrastive loss for mixed-up samples. Our proof uses the same technique as in \citep{yang_induce_nc_2022}, where they demonstrate that samples from minority class should not be learned with Cross-Entroy (CE) loss. However, in this work, we extend this proof in case of mixed-up samples when learning representation. 

\noindent\textbf{Contrastive loss.} Specifically, these samples can not be learned with current supervised contrastive loss as SupCon \citep{khosla2021supcon}. Indeed, the equation of SupCon loss is as follows,
\begin{equation}
    \mathcal{L}_{SupCon}=\sum_{i=1}^{2N}\frac{-1}{\left|P(i)\right|}\sum_{j \in P(i)}\log(\frac{e^{\langle\mathbf{z}_i \cdot \mathbf{z}_j \rangle/\tau}}{\sum_{k \in A(i)}e^{\langle\mathbf{z}_i \cdot \mathbf{z}_k\rangle/\tau}})
    \label{eq:sup-con}
\end{equation}

where $\langle\cdot\rangle$ is the cosine similarity, $\tau > 0$ is the temperature factor, $A(i) = \{1..2N\} \setminus \{i\}$, and $P(i)$ is the index set of positive views with the anchor $\mathbf{x}_i$, denoted as:

\begin{equation}
    \label{eq:set-postive-index}
    P(i) = \{p \in \{1...2N\} | y_p=y_i, p \neq i\}    
\end{equation}

 From equation of $\mathcal{L}_{SupCon}$, when using with mixed-up samples, since they have no positive views, $P(i) = \emptyset$ for all mixed-up samples. Therefore, $\mathcal{L}_{SupCon}$ is always equal to $0$. In the context of NC with fixed prototypes, recent loss functions like FNC$^2$ \citep{anh_wacv_2025} (as described in Eq. (\ref{eq:fnc2-loss})) can be used, but it is not efficient since it can break the optimal points. Similar to SupCon, when adapting $\mathcal{L}_{FNC^2}$ for mixed-up samples, $P(i) = \emptyset$ for these samples. Therefore, this loss will become as follows (for simplify, we consider this loss for a specific mixed sample $\tilde{\mathbf{x}}$ with corresponding feature $\tilde{\mathbf{z}} = g^t(f^t(\tilde{\mathbf{x}}))$) and mixed prototype $\tilde{\mathbf{p}}$,
%  \begin{equation}
%     \mathcal{L}_{FNC^2} = -\sum_{i=1}^{2N} (1-r_{i})^{\gamma} \log(r_{i})
%     \label{eq:fnc2-mix-short}
% \end{equation}
 \begin{equation}
    \mathcal{L}_{FNC^2}(\tilde{\mathbf{z}}) = - (1-\tilde{r})^{\gamma} \log(\tilde{r})
    \label{eq:fnc2-mix-short}
\end{equation}
where 
\begin{equation}
    \tilde{r}=\frac{e^{\langle\tilde{\mathbf{z}} \cdot \tilde{\mathbf{p}}\rangle/\tau}}{\sum_{{z_k} \neq \tilde{\mathbf{z}}}{e^{\langle\tilde{\mathbf{z}} \cdot \mathbf{z}_k\rangle/\tau}} + \sum_{\mathbf{p}_l \in \mathbf{P}_{1:t-1}}{e^{\langle\tilde{\mathbf{z}} \cdot \mathbf{p}_l\rangle/\tau}}}
    \label{eq:r-mix}
\end{equation}

Since the focal term $(1-\tilde{r})^{\gamma}$ is only used to encourage model focus more on hard samples which are far from their prototypes, we omit this term for simplicity and focus on the analysis:
 \begin{align}
    \mathcal{L}_{FNC^2}(\tilde{\mathbf{z}}) &= - \log(\tilde{r}) \nonumber\\
                                            &= -{\langle\tilde{\mathbf{z}} \cdot \tilde{\mathbf{p}}\rangle/\tau} + \log(A)
    \label{fnc2-mix-short-expansion}
\end{align}
where $A={\sum_{{z_k} \neq \tilde{\mathbf{z}}}{e^{\langle\tilde{\mathbf{z}} \cdot \mathbf{z}_k\rangle/\tau}} + \sum_{\mathbf{p}_l \in \mathbf{P}_{1:t-1}}{e^{\langle\tilde{\mathbf{z}} \cdot \mathbf{p}_l\rangle/\tau}}}$.

Here, the prototypes are fixed, only features are learnable, the gradient of $\mathcal{L}_{FNC^2}$ for mixed-up samples with respect to $\tilde{\mathbf{z}}$ is:

\begin{align}
    \frac{\partial\mathcal{L}_{FNC^2}}{\partial\tilde{\mathbf{z}}} &= -{\tilde{\mathbf{p}}/\tau}+\frac{1}{A}\frac{\partial A}{\partial \tilde{\mathbf{z}}} \nonumber\\
                                                                   &= -{\tilde{\mathbf{p}}/\tau} + {\sum_{\mathbf{z}_k \neq \tilde{\mathbf{z}}}{(\mathbf{z}_k/\tau)}\frac{e^{\langle\tilde{\mathbf{z}}\cdot\mathbf{z}_k\rangle}/\tau}{A}} + {(\mathbf{p}_l/\tau)}{\sum_{\mathbf{p}_l \in \mathbf{P}_{1:t-1}}\frac{e^{\langle\tilde{\mathbf{z}}\cdot\mathbf{p}_l\rangle}/\tau}{A}} \nonumber\\
                                                                   &=-{\tilde{\mathbf{p}}/\tau} + {\sum_{\mathbf{z}_k \neq \tilde{\mathbf{z}}}{(\mathbf{z}_k/\tau)}\mathcal{F}(\mathbf{z}_k)} + {\sum_{\mathbf{p}_l \in \mathbf{P}_{1:t-1}}{(\mathbf{p}_l/\tau)}{\mathcal{F}(\mathbf{p}_l)}}
    \label{eq:fnc2-gradient}
\end{align}

with $\mathcal{F}(a)=\frac{e^{\langle\tilde{\mathbf{z}}\cdot a\rangle}/\tau}{A}$.
The negative gradient $-\frac{\partial\mathcal{L}_{FNC^2}}{\partial\tilde{\mathbf{z}}}$ can be decoupled as the combination of pulling and pushing term:
\begin{equation}
    \begin{array}{l}
         \mathbf{G}_{pull} = -{\tilde{\mathbf{p}}/\tau} \vspace{1em}\\
         \mathbf{G}_{push} = {\sum_{\mathbf{z}_k \neq \tilde{\mathbf{z}}}{(\mathbf{z}_k/\tau)}\mathcal{F}(\mathbf{z}_k)} + {\sum_{\mathbf{p}_l \in \mathbf{P}_{1:t-1}}{(\mathbf{p}_l/\tau)}{\mathcal{F}(\mathbf{p}_l)}}
    \end{array}
    \label{eq:gpull-gpush}
\end{equation}
The pulling term $\mathbf{G}_{pull}$ pulls the mixed-up sample $\tilde{\mathbf{z}}$ towards its mixed prototype $\tilde{\mathbf{p}}$, while the pushing term $\mathbf{G}_{push}$ pushes $\tilde{\mathbf{z}}$ away from the other samples features $(\forall\mathbf{z}_k \neq \tilde{\mathbf{z}})$ and previous prototypes $\mathbf{P}_{1:t-1}$.  

The ``push'' gradient \( \mathbf{G}_{{push}} \) is crucial for learning classifiers. However, here, \( \tilde{\mathbf{p}} \) is not learnable as it is created by mixing fixed prototypes; therefore, the ``push'' term is unnecessary and can even cause deviation since it does not direct to the corresponding mixed prototype as the ``pull'' term does. Furthermore, it can be observed that \( \mathbf{G}_{{pull}} \) is dominated by \( \mathbf{G}_{{push}} \), making it difficult to converge and inducing a mixed-up configuration \citep{fisher2024pushingboundariesmixupsinfluence}. Building on these insights, we opt not to use contrastive loss such as \( \mathcal{L}_{{FNC}^2} \) for mixed-up samples.

% The "push" gradient $\mathbf{G}_{push}$ is crucial for learning classifiers, but here the $\tilde{\mathbf{p}}$ is not learnable, it is created by mixing fixed prototypes, so the "push" term is unnecessary.
% % Note that $\tilde{\mathbf{p}}$ is not learnable, it is created by mixing fixed prototypes. 
% Furthermore, it can be seen that $\mathbf{G}_{pull}$ is dominated by $\mathbf{G}_{push}$ making it difficult to converge and induce mixed-up configuration. Building on these insights, to avoid deviation cause by the "push" term, we do not choose contrastive loss such as $\mathcal{L}_{FNC^2}$ for mixed-up samples.

\noindent\textbf{Dot-Regression loss.}
The gradient of the Dot-Regression (DR) loss is provided in \citep{yang_induce_nc_2022}, with a mixed-up sample $\tilde{\mathbf{x}}$ and corresponding feature $\tilde{\mathbf{z}}$, it is calculated as follows,
\begin{equation}
    \frac{\partial \mathcal{L}_{DR}}{\partial \tilde{\mathbf{z}}}= - (1-\langle\tilde{\mathbf{z}} \cdot \tilde{\mathbf{p}}\rangle)\tilde{\mathbf{p}}
    \label{eq:derivative-dr}
\end{equation}
% This gradient share the same form with the first term in \cref{eq:fnc2-gradient}, which plays the role of
% “pull”, but has no “push” term.
% This gradient has the same form as the first term in \cref{eq:fnc2-gradient}, serving a "pull" "push" component.
This gradient matches the first term in Eq. (\ref{eq:fnc2-gradient}), which functions as a "pull", but it has no "push" term. Since it only pulls the mixed sample towards the corresponding mixed prototype, which is not learnable, it helps to accelerate learning without encountering the issues associated with FNC$^2$. Therefore, in this work, we choose the DR loss for learning from mixed samples.

% \subsection{Linear interpolation}
% This section show details of the squared $l2$-norm computation of the mixed-up prototype $\tilde{\mathbf{p}}_{ij}$ which is generated using linear interpolation as in \cref{eq: linear-prototypes}, as follows,

% \begin{align}
%     \|\tilde{\mathbf{p}}_{ij}\|_2^2 &= \|\lambda \mathbf{p}_{y_i} + \left(1 - \lambda\right) \mathbf{p}_{y_j} \|_2^2 \nonumber\\
%     & = \lambda^2 \|\mathbf{p}_{y_i}\|_2^2 + 2\lambda(1-\lambda) \langle\mathbf{p}_{y_i} \cdot \mathbf{p}_{y_j}\rangle \nonumber\\
%     &\qquad + {(1 - \lambda) ^ 2}{\|\mathbf{p}_{y_j}\|_2^2}
%     \label{eq:squared-l2-norm-details}
% \end{align}

% with $\Omega = \angle(\mathbf{p}_{y_i}, \mathbf{p}_{y_j})$. Since $\|\mathbf{p}_{y_i}\|_2^2 = 1$ and $\|\mathbf{p}_{y_j}\|_2^2 = 1$, \cref{eq:squared-l2-norm-details} simplifies to:
% \begin{equation}
%     \|\tilde{\mathbf{p}}_{ij}\|_2^2 = \lambda^2 + (1 - \lambda)^2 + 2{\lambda}{(1 - \lambda)}{\cos{\Omega}}
%     \label{eq:squared-l2-norm-simplify}
% \end{equation}

% \section{Neural Collapse Concept}
% \label{sec:nc-concept}

\section{Definitions of Distillation Losses}
\label{sec:def-distillation-losses}
In stability learning, we use hardness-softness distillation (HSD) due to its effectiveness compared to instance-wise relation distillation (IRD) and sample-prototype relation distillation (S-PRD). The defintion of HSD is as follows,
\begin{equation}
    \mathcal{L}_{HSD} = (1 - \xi) \mathcal{L}_{IRD} + \xi \mathcal{L}_{S-PRD}
    \label{eq:hsd}
\end{equation}
where $\xi = max(0, \frac{e - e_0}{E})$ , $e$ represents the epoch index, $e_0$ denotes the number of epochs used for the warm-up period, and $E$ is the total number of epochs.
\begin{equation}
    \left\{
    \begin{array}{l} 
        \mathcal{L}_{IRD} = \sum_{i=1}^{2N}{-\mathbf{o}^{t-1}(\mathbf{x}_i) \cdot \log(\mathbf{o}^{t}(\mathbf{x}_i))} \\
        \mathcal{L}_{S-PRD} = \sum_{i=1}^{2N} -\mathbf{q}^{t-1}(\mathbf{x}_i;\mathbf{P}_{1:t}) 
\cdot\log(\textbf{q}^t(\mathbf{x}_i;\mathbf{P}_{1:t}))
    \end{array}
    \right.
    \label{eq:detail-ird-sprd}
\end{equation}
In the loss $\mathcal{L}_{IRD}$ \citep{co2l}, $\mathbf{o}^t(\mathbf{x}_i)$ is defined as:
\begin{equation}
    \mathbf{o}^t(\mathbf{x}_i) =[o^t_{i,1},...,o^t_{i,i-1},o^t_{i},o^t_{i,i+1},...,o^t_{i,2N}]
    \label{eq:defination-oij}
\end{equation}
Each individual element $o^t_{i,1}$ is computed as:
 \begin{equation}
     o_{i,j}^t=\frac{e^{\langle\mathbf{z}_i^t \cdot \mathbf{z}_j^t\rangle/\kappa}}{\sum_{k\neq{i}}e^{\langle\mathbf{z}_i^t \cdot \mathbf{z}_k^t\rangle/\kappa}}
 \end{equation}
where $\kappa$ is the temperature hyperparameter, and $t \ge 1$ is the task index.
For the loss $\mathcal{L}_{S-PRD}$ \citep{anh_wacv_2025}, given the prototype set $\mathbf{P}_{1:t} = \{\mathbf{p}_s\}_{s=1}^{S}$, which includes all prototypes used from task $1$ to task $t$. The definition of $\mathcal{L}_{S-PRD}$ is defined as,
\begin{equation}
    \mathbf{q}^t{(\mathbf{x}_i;\mathbf{P}_{1:t})}=[q^t_{i,1}, q^t_{i,2},...,q^t_{i,S}]
\end{equation}
where each ${q}_{i,j}^t$ is computed as:
\begin{equation}
    q_{i,j}^t=\frac{e^{\langle\mathbf{z}_i^t \cdot \mathbf{p}_j\rangle/\zeta}}{\sum_{s=1}^{S}{e^{\langle\mathbf{z}_i^t \cdot \mathbf{p}_s\rangle/\zeta}}}
\end{equation}
with $\zeta$ being the temperature factor.
% each $o^t_{i,j}$ is: $$,
% where $t \ge 1$ is the task index and $\kappa$ is the temperature hyperparameter.
% In $\mathcal{L}_{{S-PRD}}$, $\mathbf{P}_{1:t} = \{\mathbf{p}_s\}_{s=1}^{S}$ is the set of prototypes used from task $1$ to task $t$.
% % , $S$ is the number of used prototypes, $S = |\mathbf{P}_{1:t}|$
% Besides, we have $\mathbf{q}^t{(\mathbf{x}_i;\mathbf{P}_{1:t})}=[q^t_{i,1}, q^t_{i,2},...,q^t_{i,S}]$, and each ${q}_{i,j}^t$ is computed as: $q_{i,j}^t=\frac{e^{\langle\mathbf{z}_i^t \cdot \mathbf{p}_j\rangle/\zeta}}{\sum_{s=1}^{S}{e^{\langle\mathbf{z}_i^t \cdot \mathbf{p}_s\rangle/\zeta}}}$, given the temperature factor $\zeta$.

\section{Additional Training Details}

\subsection{Data Preparation}
Similar to other baselines \citep{co2l, anh_wacv_2025, wen2024provablecontrastivecontinuallearning}, we apply a standard set of augmentations in the CL literature for each image as follows (using Pytorch \citep{paszke2019pytorchimperativestylehighperformance} notations): 
\begin{itemize}
    \item \texttt{RandomResizedCrop.} We apply random cropping with scales of $[0.2, 1.0]$, $[0.2, 1.0]$, $[0.1, 1.0]$ for Seq-Cifar-10, Seq-Cifar-100, and Seq-Tiny-ImageNet, respectively, and resize the images to $32 \times 32$, $32 \times 32$, and $64 \times 64$.
    \item \texttt{RandomHorizontalFlip.} We horizontally flip each image with a probability of 0.5.
    \item \texttt{ColorJitter.} We adjust brightness, contrast, saturation, and hue with maximum strengths of 0.4, 0.4, 0.4, and 0.1, respectively, with probability 0.8.
    \item \texttt{RandomGrayscale.} We convert images to grayscale with a probability of 0.2 to increase appearance invariance.
    \item \texttt{GaussianBlur.} For Seq-Tiny-ImageNet, Gaussian blur is applied with a $7 \times 7$ kernel and a standard deviation randomly sampled from $[0.1, 2.0]$, with a probability of 0.5.
\end{itemize}
In addition, following \citep{buzzega2020darkexperiencegeneralcontinual, co2l, anh_wacv_2025, wen2024provablecontrastivecontinuallearning}, each batch is sampled equally and independently from the union of current task data and memory, eliminating the need to control old samples, avoiding sample imbalances and single-class batches that can degrade contrastive loss like FNC$^2$ \citep{anh_wacv_2025}.
% Similar to prior CL work \citep{buzzega2020darkexperiencegeneralcontinual, co2l, wen2024provablecontrastivecontinuallearning, anh_wacv_2025}, our method uses a standard set of augmentations as follows:

\subsection{Hyperparameter Selection}
% Note that, to enable comparison, we reproduce Co$^2$L \citep{co2l} on Seq-Cifar-100 and the memory-free setting of Seq-Tiny-ImageNet, as these results are not reported in the original paper. 
We utilize a grid search method to select the optimal hyperparameters, using a randomly drawn $10\%$  of the training data as the validation set. 
% Selection is based on the average validation accuracy across 5 independent trials with different random seeds.
\begin{table}[t]
    \caption{Search spaces of all hyperparameters. Baselines use the top set; while SAMix tunes \textbf{only 4} additional hyperparameters below}
    \centering
    \setlength{\tabcolsep}{9mm}
    % \resizebox*{!}{0.84\columnwidth}{
        % \small
        % \input{AnonymousSubmission/LaTeX/tables/table_search_space_hyperparameters}
        \begin{tabular}{cc}
            \toprule
            \textbf{Hyperparameter} & \textbf{Values} \\
            \midrule
            $E_1$  & \{500\} \\
            $E_{t \geq 2}$ & \{50, 100\} \\
            $\eta$ & \{0.1, 0.5, 1.0\}\\
            \textit{bsz} & \{256, 512\} \\
            $\gamma$ & \{0, 1, 2, 4, 7, 10\} \\
            $\tau$ & \{0.1, 0.5, 1.0\} \\
            $\kappa_{past}$ & \{0.01, 0.05, 0.1\} \\
            $\kappa_{current}$ & \{0.1, 0.2\} \\
            $\zeta_{past}$ & \{0.01, 0.05, 0.1\} \\
            $\zeta_{current}$ & \{0.1, 0.2\} \\
            $e_0$ & \{10, 20, 30\} \\
            \midrule
            $\alpha$ & \{1, 10, 25\} \\
            $M$ & \{5, 10, 15\} \\
            $\upsilon$ & \{1, 2, 5, 10\} \\
            $\iota$ & \{1, 2, 5, 10\} \\
            \bottomrule
        \end{tabular}
    % }
    \label{tab:search-space-hyperparameters}
\end{table}

\noindent\textbf{Baseline.} The baseline hyperparameters under evaluation include:

\begin{itemize}
    \item Number of initial epochs ($E_1$)
    \item Number of epochs of $t$-th task ($E_{t \ge 2}$)
    \item Learning rate ($\eta$)
    \item Batch size (\textit{bsz})
    \item Focusing hyperparameters ($\gamma$) for FNC$^2$ loss
    \item Temperature for the Focal Neural Collapse Contrastive (FNC$^2$) ($\tau$)
    \item Temperature for instance-wise relation distillation loss ($\mathcal{L}_{IRD}$): Following \citep{co2l}, we employ distinct temperature hyperparameters for the similarity vectors of past ($\kappa_{past}$) and current ($\kappa_{current}$) data
    \item Temperature for sample-prototype relation distillation loss ($\mathcal{L}_{S-PRD}$): We use different temperature hyperparameters, $\zeta_{past}$ for past similarity vectors and $\zeta_{current}$ for current similarity vectors
    \item Number of warm-up epochs in hardness-softness distillation loss ($\mathcal{L}_{HSD}$) ($e_0$)
    % \text{Beta}(\alpha, \alpha)$, $\alpha \in (0, \infty)$
\end{itemize}

These baseline hyperparameters are \textbf{tuned once} for the baselines \textit{without SAMix} and \textbf{kept fixed} \textit{when integrating SAMix}. 

\noindent\textbf{When using SAMix.} In addition to these tuned baseline values, we further tune only four hyperparameters: $M$, $\alpha$, $\upsilon$, and $\iota$, listed as:

\begin{itemize}
    \item Hyperparameter ($\alpha$) of the $\text{Beta}(\alpha, \alpha)$ to control the mixing coefficient $\lambda$ in SAMix
    \item Number of bins (M) to compute the expected calibration error
    \item Balancing hyperparameters $\upsilon$ and $\iota$ for learning mixed samples in the DR-based and FNC$^2$-based plasticity methods, respectively
\end{itemize}

\noindent The search space for all hyperparameters is listed in Table \ref{tab:search-space-hyperparameters}. 
% Selection is based on the average test accuracy over 5 independent trials with 5 different seeds. 
The final chosen values for the baselines and SAMix-integrated methods are detailed in Table \ref{tab:chosen-baseline-hyperparameters} and Table \ref{tab:chosen-samix-hyperparameters}, respectively. When \textbf{using SAMix}, we select $M=15$, $\alpha = 25$ for all scenarios, and $\upsilon = \iota = 5$ for both the FC-NCCL and TA-NCCL methods. 
% In addition, we use seed number 5 for all experiments.  
To maintain focus, we exclude hyperparameters previously discovered in the literature.

\subsection{Training Schedules}
\textbf{In training. } For all experiments, we choose SGD as the optimizer with momentum of 0.9 and weight decay of 0.0001. The learning rate is warmed up for 10 epochs and then decayed using the cosine schedule \citep{loshchilov2017sgdrstochasticgradientdescent}, which restarts at the beginning of each new task.

\noindent\textbf{In evaluation.} A linear classifier is trained for 100 epochs using SGD with momentum 0.9 and no weight decay. The initial learning rates are {1.0, 0.1, 0.1} for Seq-Cifar-10, Seq-Cifar-100, Seq-Tiny-ImageNet, respectively, then decayed exponentially by a factor of 0.2 at epochs 60, 75, and 90.

\begin{table*}[ht]
    \caption{Selected hyperparameters when integrating SAMix. Differences from the baselines are shown in bold, \textbf{\textit{only 4}} additional hyperparameters are tuned}
    \centering
    \resizebox*{!}{0.6\linewidth}{
    \small
    \begin{tabular}{cccc}
        \toprule
        \textbf{Method} & \textbf{Buffer size} & \textbf{Dataset} & \textbf{Hyperparameter} \\
        \midrule
        \multirow{12}{*}{\makecell{FC-NCCL \\+ SAMix}}
        
        &\multirow{4}{*}{0, 200, 500}&\multirow{3}{*}{Seq-Cifar-10}&\textit{bsz}: 512, $\eta$: 0.5, $\gamma$: 1, $E_1$: 500, $E_{t \ge 2}$: 100, \\
        &&&$\tau$: 0.5, $e_0$: 30, $\kappa_{past}$: 0.01, $\kappa_{current}$: 0.2, \\
        &&&$\zeta_{past}$: 0.01, $\zeta_{current}$: 0.2, \\
        % &&&${M: 15}$, ${\alpha : 25}$, $\upsilon: 5$, $\iota: 5$ \\
        &&&\textbf{${\mathbf{M}: \mathbf{15}}$, ${\boldsymbol{\alpha} : \mathbf{25}}$, ${\boldsymbol{\upsilon}: \mathbf{5}}$, ${\boldsymbol{\iota}: \mathbf{5}}$} \\
        
        \cmidrule{2-4} 
        
        & \multirow{4}{*}{0, 200, 500} & \multirow{3}{*}{Seq-Cifar-100}&\textit{bsz}: 512, $\eta$: 0.5, $\gamma$: 4, $E_1$: 500, $E_{t \ge 2}$: 100, \\
        &&&$\tau$: 0.5, $e_0$: 30, $\kappa_{past}$: 0.01, $\kappa_{current}$: 0.2, \\
        &&&$\zeta_{past}$: 0.1, $\zeta_{current}$: 0.2, \\
        &&&\textbf{${\mathbf{M}: \mathbf{15}}$, ${\boldsymbol{\alpha} : \mathbf{25}}$, ${\boldsymbol{\upsilon}: \mathbf{5}}$, ${\boldsymbol{\iota}: \mathbf{5}}$} \\
    
        \cmidrule{2-4} 
        
        & \multirow{4}{*}{0, 200, 500} & \multirow{3}{*}{Seq-Tiny-ImageNet}&\textit{bsz}: 512, $\eta$: 0.1, $\gamma$: 4, $E_1$: 500, $E_{t \ge 2}$: 50, \\
        &&&$\tau$: 0.5, $e_0$: 20, $\kappa_{past}$: 0.1, $\kappa_{current}$: 0.1, \\
        &&&$\zeta_{past}$: 0.1, $\zeta_{current}$: 0.2, \\
        &&&\textbf{${\mathbf{M}: \mathbf{15}}$, ${\boldsymbol{\alpha} : \mathbf{25}}$, ${\boldsymbol{\upsilon}: \mathbf{5}}$, ${\boldsymbol{\iota}: \mathbf{5}}$} \\

        \midrule
        \multirow{12}{*}{\makecell{TA-NCCL \\+ SAMix}}
        &\multirow{4}{*}{0, 200, 500}&\multirow{3}{*}{Seq-Cifar-10}&\textit{bsz}: 512, $\eta$: 0.5, $E_1$: 500, $E_{t \ge 2}$: 100, \\
        &&&$\tau$: 0.5, $e_0$: 30, $\kappa_{past}$: 0.01, $\kappa_{current}$: 0.2,\\
        &&&$\zeta_{past}$: 0.01, $\zeta_{current}$: 0.2, \\
        &&&\textbf{${\mathbf{M}: \mathbf{15}}$, ${\boldsymbol{\alpha} : \mathbf{25}}$, ${\boldsymbol{\upsilon}: \mathbf{5}}$, ${\boldsymbol{\iota}: \mathbf{5}}$} \\
        
        \cmidrule{2-4} 
        
        & \multirow{4}{*}{0, 200, 500} & \multirow{3}{*}{Seq-Cifar-100}&\textit{bsz}: 512, $\eta$: 0.5, $E_1$: 500, $E_{t \ge 2}$: 100, \\
        &&&$\tau$: 0.5, $e_0$: 30, $\kappa_{past}$: 0.01, $\kappa_{current}$: 0.2, \\
        &&&$\zeta_{past}$: 0.1, $\zeta_{current}$: 0.2, \\
        &&&\textbf{${\mathbf{M}: \mathbf{15}}$, ${\boldsymbol{\alpha} : \mathbf{25}}$, ${\boldsymbol{\upsilon}: \mathbf{5}}$, ${\boldsymbol{\iota}: \mathbf{5}}$} \\
        
        \cmidrule{2-4} 
        
        & \multirow{3}{*}{0, 200, 500} & \multirow{3}{*}{Seq-Tiny-ImageNet}&\textit{bsz}: 512, $\eta$: 0.1, $E_1$: 500, $E_{t \ge 2}$: 50, \\
        &&&$\tau$: 0.5, $e_0$: 20, $\kappa_{past}$: 0.1, $\kappa_{current}$: 0.1, \\
        &&&$\zeta_{past}$: 0.1, $\zeta_{current}$: 0.2, \\
        &&&\textbf{${\mathbf{M}: \mathbf{15}}$, ${\boldsymbol{\alpha} : \mathbf{25}}$, ${\boldsymbol{\upsilon}: \mathbf{5}}$, ${\boldsymbol{\iota}: \mathbf{5}}$} \\
        \bottomrule
    \end{tabular}
    }
    \label{tab:chosen-samix-hyperparameters}
\end{table*}

\begin{table*}[ht]
    \caption{Selected baseline hyperparameters in our experiments}
    \centering
    \resizebox*{!}{0.5\textwidth}{
    \small
    \begin{tabular}{cccc}
        \toprule
        \textbf{Method} & \textbf{Buffer size} & \textbf{Dataset} & \textbf{Hyperparameter} \\
        \midrule
        \multirow{9}{*}{FC-NCCL}
        
        &\multirow{3}{*}{0, 200, 500}&\multirow{3}{*}{Seq-Cifar-10}&\textit{bsz}: 512, $\eta$: 0.5, $\gamma$: 1, $E_1$: 500, $E_{t \ge 2}$: 100, \\
        &&&$\tau$: 0.5, $e_0$: 30, $\kappa_{past}$: 0.01, $\kappa_{current}$: 0.2,\\
        &&&$\zeta_{past}$: 0.01, $\zeta_{current}$: 0.2 \\
        
        \cmidrule{2-4} 
        
        & \multirow{3}{*}{0, 200, 500} & \multirow{3}{*}{Seq-Cifar-100}&\textit{bsz}: 512, $\eta$: 0.5, $\gamma$: 4, $E_1$: 500, $E_{t \ge 2}$: 100, \\
        &&&$\tau$: 0.5, $e_0$: 30, $\kappa_{past}$: 0.01, $\kappa_{current}$: 0.2, \\
        &&&$\zeta_{past}$: 0.1, $\zeta_{current}$: 0.2 \\
    
        \cmidrule{2-4} 
        
        & \multirow{3}{*}{0, 200, 500} & \multirow{3}{*}{Seq-Tiny-ImageNet}&\textit{bsz}: 512, $\eta$: 0.1, $\gamma$: 4, $E_1$: 500, $E_{t \ge 2}$: 50, \\
        &&&$\tau$: 0.5, $e_0$: 20, $\kappa_{past}$: 0.1, $\kappa_{current}$: 0.1, \\
        &&&$\zeta_{past}$: 0.1, $\zeta_{current}$: 0.2 \\

        \midrule
        \multirow{9}{*}{TA-NCCL}
        &\multirow{3}{*}{0, 200, 500}&\multirow{3}{*}{Seq-Cifar-10}&\textit{bsz}: 512, $\eta$: 0.5, $E_1$: 500, $E_{t \ge 2}$: 100, \\
        &&&$\tau$: 0.5, $e_0$: 30, $\kappa_{past}$: 0.01, $\kappa_{current}$: 0.2,\\
        &&&$\zeta_{past}$: 0.01, $\zeta_{current}$: 0.2 \\
        
        \cmidrule{2-4} 
        
        & \multirow{3}{*}{0, 200, 500} & \multirow{3}{*}{Seq-Cifar-100}&\textit{bsz}: 512, $\eta$: 0.5, $E_1$: 500, $E_{t \ge 2}$: 100, \\
        &&&$\tau$: 0.5, $e_0$: 30, $\kappa_{past}$: 0.01, $\kappa_{current}$: 0.2, \\
        &&&$\zeta_{past}$: 0.1, $\zeta_{current}$: 0.2 \\
    
        \cmidrule{2-4} 
        
        & \multirow{3}{*}{0, 200, 500} & \multirow{3}{*}{Seq-Tiny-ImageNet}&\textit{bsz}: 512, $\eta$: 0.1, $E_1$: 500, $E_{t \ge 2}$: 50, \\
        &&&$\tau$: 0.5, $e_0$: 20, $\kappa_{past}$: 0.1, $\kappa_{current}$: 0.1, \\
        &&&$\zeta_{past}$: 0.1, $\zeta_{current}$: 0.2 \\
        \bottomrule
    \end{tabular}
    }
    \label{tab:chosen-baseline-hyperparameters}
\end{table*}

\section{Additional Results}

\subsection{Average forgetting results}
\label{sec:average-forgetting}
The Average Forgetting metric \citep{chaudhry_2018}, used in this work to quantify model forgetting, is defined as: 
\begin{equation}
    F = \frac{1}{T-1}\sum_{i=1}^{T-1}{max_{t \in \{1,\ldots,T-1\}}{(A_{t,i}-A_{T,i})}}
    \label{eq:average-forgetting}
\end{equation}
Table \ref{tab:forgetting-result} reports the average forgetting results of our method compared to all other baselines. The results demonstrate that our method effectively mitigates forgetting, even without relying on additional buffers.

\begin{table*}[t]
    \caption{Average forgetting (lower is better) across 5 independent trials from 5 different seeds: Comparing our method against all baselines in continual learning (\textbf{\text{Ours: TA-NCCL}} method)}
    \centering
    \small
    \setlength{\tabcolsep}{1.2mm}  %1.6pt \renewcommand{\arraystretch}{1.2}
    \renewcommand{\arraystretch}{1.1}
    % \resizebox{1.0\linewidth}{!}{
    \resizebox{0.98\linewidth}{!}{
        \begin{tabular}{clccccccc}
        \hline
        \multirow{2}{*}{\textbf{Buffer}} & \textbf{Dataset} & \multicolumn{2}{c}{\textbf{Seq-Cifar-100}} & \multicolumn{2}{c}{\textbf{Seq-Tiny-ImageNet}} & \multicolumn{2}{c}{\textbf{Seq-Cifar-10}}\\
                     & \textbf{Scenario} & \textbf{Class-IL} & \textbf{Task-IL} & \textbf{Class-IL} & \textbf{Task-IL} & \textbf{Class-IL} & \textbf{Task-IL}\\
        \hline
            \multirow{5}{*}{0}
            &{Co$^2$L} (ICCV'21) \citep{co2l} &66.51$\pm$0.28&39.63$\pm$0.62&62.80$\pm$0.77&39.54$\pm$1.08&35.81$\pm$1.08&14.33$\pm$0.87\\
            \cmidrule{2-8}
            & \grayrow{FC-NCCL (WACV'25) \citep{anh_wacv_2025}} {52.03$\pm$0.63} {36.20$\pm$0.48} {53.97$\pm$0.63} {37.57$\pm$0.88}{23.85$\pm$0.30} {4.72$\pm$0.28}\\
            & \grayrow{FC-NCCL + SAMix} {44.30$\pm$0.59} {26.54$\pm$0.88} {52.39$\pm$0.75} {37.16$\pm$0.53} {27.45$\pm$1.49} {7.47$\pm$1.22}\\
            & \grayrow{Ours} {44.69$\pm$0.67} {\textbf{19.07$\pm$0.73}} {\textbf{41.38$\pm$0.89}} {\textbf{27.99$\pm$0.76}} {22.20$\pm$0.79} {2.05$\pm$0.18}\\
            & \grayrow{Ours + SAMix} {\textbf{42.53$\pm$0.97}} {20.69$\pm$0.75} {51.61$\pm$0.55} {36.53$\pm$0.62} {\textbf{21.47$\pm$0.68}} {\textbf{2.01$\pm$0.29}}\\
        \hline
            \multirow{12}{*}{200}
            &ER (ICLR'19) \citep{riemer2019learninglearnforgettingmaximizing} &75.06$\pm$0.63&	27.38$\pm$1.46	&76.53$\pm$0.51&	40.47$\pm$1.54 &59.30$\pm$2.48&6.07$\pm$1.09
             \\
            % &GEM \citep{lopezpaz2017gem}&77.40$\pm$1.09&29.59$\pm$1.66
            % &-&- &80.36$\pm$5.25&9.57$\pm$2.05\\ 
            % &GSS \citep{aljundi2019gradientbasedsampleselection}&	77.62$\pm$0.76&	32.81$\pm$1.75&	- &	- &72.48$\pm$4.45&	8.49$\pm$2.05
            %  \\
            % &iCARL \citep{rebuffi2017icarl} & CVPR'17 & {47.20$\pm$1.23}&36.20$\pm$1.85&	\textbf{31.06$\pm$1.91}&	42.47$\pm$2.47 &{23.52$\pm$1.27}&	25.34$\pm$1.64
            %  \\
            &DER (NeurIPS'20) \citep{buzzega2020darkexperiencegeneralcontinual} &62.72$\pm$2.69&25.98$\pm$1.55 &64.83$\pm$1.48&40.43$\pm$1.05 &35.79$\pm$2.59& 6.08$\pm$0.70\\
            &Co$^2$L (ICCV'21) \citep{co2l} &67.82$\pm$0.41&38.22$\pm$0.34&73.25$\pm$0.21&47.11$\pm$1.04 &36.35$\pm$1.16& 6.71$\pm$0.35
            \\
            &GCR (CVPR'22) \citep{tiwari2022gcrgradientcoresetbased} &57.65$\pm$2.48&{24.12$\pm$1.17}&65.29$\pm$1.73&40.36$\pm$1.08 &32.75$\pm$2.67& 7.38$\pm$1.02\\
        
            &{CILA} (ICML'24) \citep{wen2024provablecontrastivecontinuallearning} &-&-&-&-&-&-\\
            &{CCLIS} (AAAI'24) \citep{li_cl_important_sampling_2024} & 46.89$\pm$0.59&14.17$\pm$0.20&62.21$\pm$0.34&33.20$\pm$0.75&22.59$\pm$0.18&2.08$\pm$0.27\\
            \cmidrule{2-8}
            & \grayrow{FC-NCCL (WACV'25) \citep{anh_wacv_2025}} {52.40$\pm$0.83} {33.66$\pm$0.24} {52.07$\pm$0.46} {{33.76$\pm$0.58}} {25.24$\pm$0.69} {{4.28$\pm$0.32}}\\
            & \grayrow{FC-NCCL + SAMix} {42.83$\pm$0.17} {25.29$\pm$1.03} {50.37$\pm$0.33} {32.77$\pm$0.46} {26.64$\pm$0.88} {4.34$\pm$0.25}\\
            & \grayrow{Ours} {42.40$\pm$0.43} {\textbf{13.59$\pm$1.51}} {\textbf{42.22$\pm$0.44}} {\textbf{27.85$\pm$0.58}} {20.89$\pm$1.05} {1.84$\pm$0.30}\\
            & \grayrow{Ours + SAMix} {\textbf{39.24$\pm$1.26}} {{16.51$\pm$0.53}} {51.66$\pm$0.75} {30.89$\pm$0.53} {\textbf{20.54$\pm$0.44}} {\textbf{1.27$\pm$0.21}}\\
        \hline
            \multirow{12}{*}{500}
            &ER (ICLR'19) \citep{riemer2019learninglearnforgettingmaximizing} &67.96$\pm$0.78&	17.37$\pm$1.06	&75.21$\pm$0.54&	30.73$\pm$0.62 & 43.22$\pm$2.10&3.50$\pm$0.53
             \\
            % &GEM \citep{lopezpaz2017gem}&71.34$\pm$0.78&	20.44$\pm$1.13& - & - &78.93$\pm$6.53&5.60$\pm$0.96
            %  \\
            % &GSS \citep{aljundi2019gradientbasedsampleselection}&	74.12$\pm$0.42&	26.57$\pm$1.34&	- & -&59.18$\pm$4.00&	6.37$\pm$1.55
            %  \\
            % &iCARL \citep{rebuffi2017icarl} & CVPR'17 &	40.99$\pm$1.02&	27.90$\pm$1.37	&{37.30$\pm$1.42}	&39.44$\pm$0.84&28.20$\pm$2.41&	22.61$\pm$3.97
            %  \\
            &DER (NeurIPS'20) \citep{buzzega2020darkexperiencegeneralcontinual} &49.07$\pm$2.54& 25.98$\pm$1.55&59.95$\pm$2.31&28.21$\pm$0.97&24.02$\pm$1.63&3.72$\pm$0.55\\
            
            &Co$^2$L (ICCV'21) \citep{co2l}
             &51.23$\pm$0.65&26.30$\pm$0.57
             &65.15$\pm$0.26&39.22$\pm$0.69 &25.33$\pm$0.99&3.41$\pm$0.80
            \\
            
            &GCR (CVPR'22) \citep{tiwari2022gcrgradientcoresetbased} &{39.20$\pm$2.84}& {15.07$\pm$1.88}&56.40$\pm$1.08&27.88$\pm$1.19&{19.27$\pm$1.48}&{3.14$\pm$0.36}\\
        
            &{CILA} (ICML'24) \citep{wen2024provablecontrastivecontinuallearning} &-&-&-&-&-&-\\
            &{CCLIS} (AAAI'24) \citep{li_cl_important_sampling_2024} &42.53$\pm$0.64&12.68$\pm$1.33&50.15$\pm$0.20&23.46$\pm$0.93&18.93$\pm$0.61&1.69$\pm$0.12\\
            \cmidrule{2-8}
            & \grayrow{FC-NCCL (WACV'25) \citep{anh_wacv_2025}} {41.66$\pm$0.78} {24.84$\pm$0.91} {46.08$\pm$0.56} {{26.45$\pm$0.79}} {22.59$\pm$1.02} {3.21$\pm$0.25}\\
            & \grayrow{FC-NCCL + SAMix} {34.23$\pm$0.14} {17.26$\pm$0.80} {45.14$\pm$0.76} {25.31$\pm$0.84} {20.07$\pm$0.72} {3.94$\pm$0.25}\\
            & \grayrow{Ours} {31.38$\pm$1.48} {\textbf{9.66$\pm$0.79}} {\textbf{35.62$\pm$0.62}} {\textbf{21.51$\pm$0.93}} {16.85$\pm$0.43} {1.39$\pm$0.20}\\
            & \grayrow{Ours + SAMix} {\textbf{29.94$\pm$1.06      }} {9.98$\pm$0.91} {43.53$\pm$0.59} {22.87$\pm$0.61} {\textbf{15.25$\pm$0.53}} {\textbf{1.16$\pm$0.12}}\\
        \hline
        \end{tabular}
    }
    \label{tab:forgetting-result}
\end{table*}

\subsection{Calibration Error and Overconfidence}
% ---------------------AECE-AOE TABLE----------------------------
\begin{table*}[tb]
    \caption{AECE, AOE, and last task accuracy $(A_{T,T})$ across different NC-based methods. \text{$\blacktriangle$}/\text{$\triangledown$} indicate methods applying SAMix outperforms/underperforms compared to methods without SAMix. All results indicate that SAMix is highly effective on large, complex datasets such as Seq-Cifar-100 and Seq-Tiny-ImageNet but largely ineffective on smaller datasets like Seq-Cifar-10 
    % \textbf{This outcome aligns perfectly with our hypotheses in Sec. \ref{sec:expected-outcome-hypothesis}}.
    }
    \centering
    
    % \small
    \setlength{\tabcolsep}{1.2mm}  %1.6pt \renewcommand{\arraystretch}{1.2}
    \renewcommand{\arraystretch}{1.1}
    \resizebox{!}{0.15\linewidth}{
        \begin{tabular}{clllllllllll}
        \hline
        
        \multirow{2}{*}{\textbf{Buffer}}&\textbf{Dataset}&\multicolumn{3}{c}{\textbf{Seq-Cifar-100}}&\multicolumn{3}{c}{\textbf{Seq-Tiny-ImageNet}}&\multicolumn{3}{c}{\textbf{Seq-Cifar-10}}\\
                    &\textbf{Scenario}&\textbf{AECE($\downarrow$)}&\textbf{AOE($\downarrow$)}&\textbf{A$_{T,T}$($\uparrow$)}&\textbf{AECE($\downarrow$)}&\textbf{AOE($\downarrow$)}&\textbf{A$_{T,T}$($\uparrow$)}&\textbf{AECE($\downarrow$)}&\textbf{AOE($\downarrow$)}&\textbf{A$_{T,T}$($\uparrow$)}\\
        \hline
            \multirow{4}{*}{0}
            % &Co$^2$L \citep{co2l}&58.892.61&86.651.05&26.890.78&51.910.63&13.430.57&40.210.68\\
            &FC-NCCL \citep{anh_wacv_2025}& 0.160 & 0.107 & 74.25 & 0.130 & 0.050 & 69.70 & 0.206 & 0.167 & 88.30 \\
            &FC-NCCL + SAMix & 0.141 \text{$\blacktriangle$} & 0.082 \text{$\blacktriangle$}& 77.00 \text{$\blacktriangle$}& 0.088 \text{$\blacktriangle$}& 0.021 \text{$\blacktriangle$}& 74.40 \text{$\blacktriangle$}& 0.177 \text{$\blacktriangle$}& 0.133 \text{$\blacktriangle$}& 85.60 \text{$\triangledown$}\\
            &Ours& 0.133 & 0.029 & 46.60 & 0.258 & 0.117 & 57.10 & 0.098 & 0.055 & 80.20\\
            &Ours + SAMix & 0.106 \text{$\blacktriangle$} & 0.005 \text{$\blacktriangle$}& 78.55 \text{$\blacktriangle$}& 0.146 \text{$\blacktriangle$} & 0.069 \text{$\blacktriangle$}& 63.30 \text{$\blacktriangle$}& 0.152 \text{$\triangledown$}& 0.099 \text{$\triangledown$} & 85.15 \text{$\blacktriangle$}\\
        \hline
            \multirow{4}{*}{200}
            % &Co$^2$L \citep{co2l}&65.571.37&93.430.78&27.380.85&53.940.76&13.880.40&42.370.74\\
            % &{CCLIS} \citep{li_cl_important_sampling_2024}&{74.950.61}&{96.200.26}&{42.390.37}&{72.930.46}&{16.130.19}&{48.290.78}\\ 
            &FC-NCCL \citep{anh_wacv_2025}& 0.162 & 0.109 & 73.85 & 0.170 & 0.073 & 69.70 & 0.305 & 0.252 & 95.70\\
            &FC-NCCL + SAMix & 0.139 \text{$\blacktriangle$}& 0.046 \text{$\blacktriangle$} & 77.30 \text{$\blacktriangle$} & 0.095 \text{$\blacktriangle$} & 0.033 \text{$\blacktriangle$} & 73.40  \text{$\blacktriangle$}& 0.312 \text{$\triangledown$}& 0.273 \text{$\triangledown$}& 89.30 \text{$\triangledown$}\\
            &Ours & 0.140 & 0.035 & 53.80 & 0.251 & 0.112 & 59.20 &0.153&0.108&84.40\\
            &Ours + SAMix& 0.098 \text{$\blacktriangle$}& 0.017  \text{$\blacktriangle$}& 80.15  \text{$\blacktriangle$}& 0.146  \text{$\blacktriangle$}& 0.069  \text{$\blacktriangle$}& 64.70  \text{$\blacktriangle$}& 0.219 \text{$\triangledown$}& 0.172 \text{$\triangledown$}& 91.25 \text{$\blacktriangle$}\\
        \hline
            \multirow{4}{*}{500}
            % &Co$^2$L \citep{co2l}&74.260.77&95.900.26&37.020.76&62.440.36&20.120.42&53.040.69\\
            % &{CCLIS} \citep{li_cl_important_sampling_2024}&{78.570.25}&{96.180.43}&{46.080.67}&{74.510.38}&{22.880.40}&{57.040.43}\\ 
            &FC-NCCL \citep{anh_wacv_2025}& 0.231 & 0.164 & 71.95 & 0.098 & 0.040 & 63.20 & 0.334 & 0.290 & 96.40\\
            &FC-NCCL + SAMix & 0.132 \text{$\blacktriangle$}& 0.092  \text{$\blacktriangle$}& 76.50  \text{$\blacktriangle$}& 0.079  \text{$\blacktriangle$}& 0.031  \text{$\blacktriangle$} & 64.20  \text{$\blacktriangle$}& 0.267  \text{$\blacktriangle$}& 0.226  \text{$\blacktriangle$} & 93.60 \text{$\triangledown$}\\
            &Ours& 0.098 & 0.036 & 51.15 & 0.206 & 0.095 & 48.00 & 0.086 & 0.056 & 83.55 \\
            &Ours + SAMix & 0.078 \text{$\blacktriangle$}& 0.015  \text{$\blacktriangle$} & 80.05  \text{$\blacktriangle$} & 0.115  \text{$\blacktriangle$} & 0.052  \text{$\blacktriangle$} & 56.80  \text{$\blacktriangle$} & 0.150 \text{$\triangledown$}& 0.122 \text{$\triangledown$}& 92.55  \text{$\blacktriangle$}\\
        \hline
        \end{tabular}
    }

    \label{tab:ece_oe_a5_results}
\end{table*}
% ---------------------------------------------------------------
Table \ref{tab:ece_oe_a5_results} shows the changes Average Expected Calibration Error (AECE) and Average Overconfidence Error (AOE), and accuracy on the last task $(A_{T,T})$ when learning using SAMix. As can be seen from this table, NC-based methods using SAMix improve calibration and reduce overconfidence while maintaining superior performance compared to their original counterparts on complex datasets such as Seq-Cifar-10 and Seq-Tiny-ImageNet. However, on the smaller dataset Seq-Cifar-10, SAMix exhibits the opposite trend, aligning with the performance variations across tasks, as shown in Table \ref{tab:acc_result}.




\end{appendices}

%%===============Rebuttal===============
% \include{rebuttal/response}

% \FloatBarrier    % stop any leftover figures/tables from floating forward
\clearpage       % force a fresh page and flush floats

\bibliography{sn-bibliography}% common bib file

\end{document}